\documentclass[journal]{IEEEtran}

\usepackage{cite}
\usepackage{amsmath,amssymb,amsfonts}
\usepackage{algorithmic}
\usepackage{graphicx}
\usepackage{textcomp}
\usepackage{xcolor}
\usepackage{times}
\usepackage{soul}
\usepackage[utf8]{inputenc}
\usepackage{amsmath}
\usepackage{amsfonts}
\usepackage{multirow}
\usepackage{subfigure}
\usepackage{algorithm}  
\def\BibTeX{{\rm B\kern-.05em{\sc i\kern-.025em b}\kern-.08em
		T\kern-.1667em\lower.7ex\hbox{E}\kern-.125emX}}

\begin{document}

\title{Complementary Attributes: A New Clue to Zero-Shot Learning}

\author{Xiaofeng Xu,
        Ivor W. Tsang
        and~Chuancai Liu
        \thanks{X. Xu is with the School of Computer Science and Engineering, Nanjing University of Science and Technology, Nanjing 210094, China, and also with the Centre for Artificial Intelligence, University of Technology Sydney, Ultimo NSW 2007, Australia (email: csxuxiaofeng@njust.edu.cn).}
        \thanks{I. W. Tsang is with the Centre for Artificial Intelligence, University of Technology Sydney, Ultimo NSW 2007, Australia (e-mail: ivor.tsang@uts.edu.au).}
        \thanks{C. Liu is with the School of Computer Science and Engineering, Nanjing University of Science and Technology, Nanjing 210094, China, and also with the Collaborative Innovation Center of IoT Technology and Intelligent Systems, Minjiang University, Fuzhou 350108, China (e-mail: chuancailiu@njust.edu.cn).}
        \thanks{This work was supported in part by ARC under Grant LP150100671 and Grant DP180100106, in part by NSFC under Grant 61373063 and Grant 61872188, in part by the Project of MIIT under Grant E0310/1112/02-1, in part by the Collaborative Innovation Center of IoT Technology and Intelligent Systems of Minjiang University under Grant IIC1701, and in part by China Scholarship Council. 
        \textit{(Corresponding author: Chuancai Liu.)}
		}
}

\markboth{IEEE TRANSACTIONS ON CYBERNETICS}%
{Shell \MakeLowercase{\textit{et al.}}: Bare Demo of IEEEtran.cls for IEEE Journals}

\maketitle

\begin{abstract}
Zero-shot learning (ZSL) aims to recognize unseen objects using disjoint seen objects via sharing attributes. The generalization performance of ZSL is governed by the attributes, which transfer semantic information from seen classes to unseen classes. To take full advantage of the knowledge transferred by attributes, in this paper, we introduce the notion of complementary attributes (CA), as a supplement to the original attributes, to enhance the semantic representation ability. Theoretical analyses demonstrate that complementary attributes can improve the PAC-style generalization bound of original ZSL model. Since the proposed CA focuses on enhancing the semantic representation, CA can be easily applied to any existing attribute-based ZSL methods, including the label-embedding strategy based ZSL (LEZSL) and the probability-prediction strategy based ZSL (PPZSL). In PPZSL, there is a strong assumption that all the attributes are independent of each other, which is arguably unrealistic in practice. To solve this problem, a novel rank aggregation framework is proposed to circumvent the assumption. Extensive experiments on five ZSL benchmark datasets and the large-scale ImageNet dataset demonstrate that the proposed complementary attributes and rank aggregation can significantly and robustly improve existing ZSL methods and achieve the state-of-the-art performance.
\end{abstract}


\IEEEpeerreviewmaketitle

\section{Introduction}

\begin{figure}[t]
	\centering
	\subfigure[Label-Embedding Strategy]{
		\label{fig_zslframe_1}
		\includegraphics[width = 0.43\textwidth]{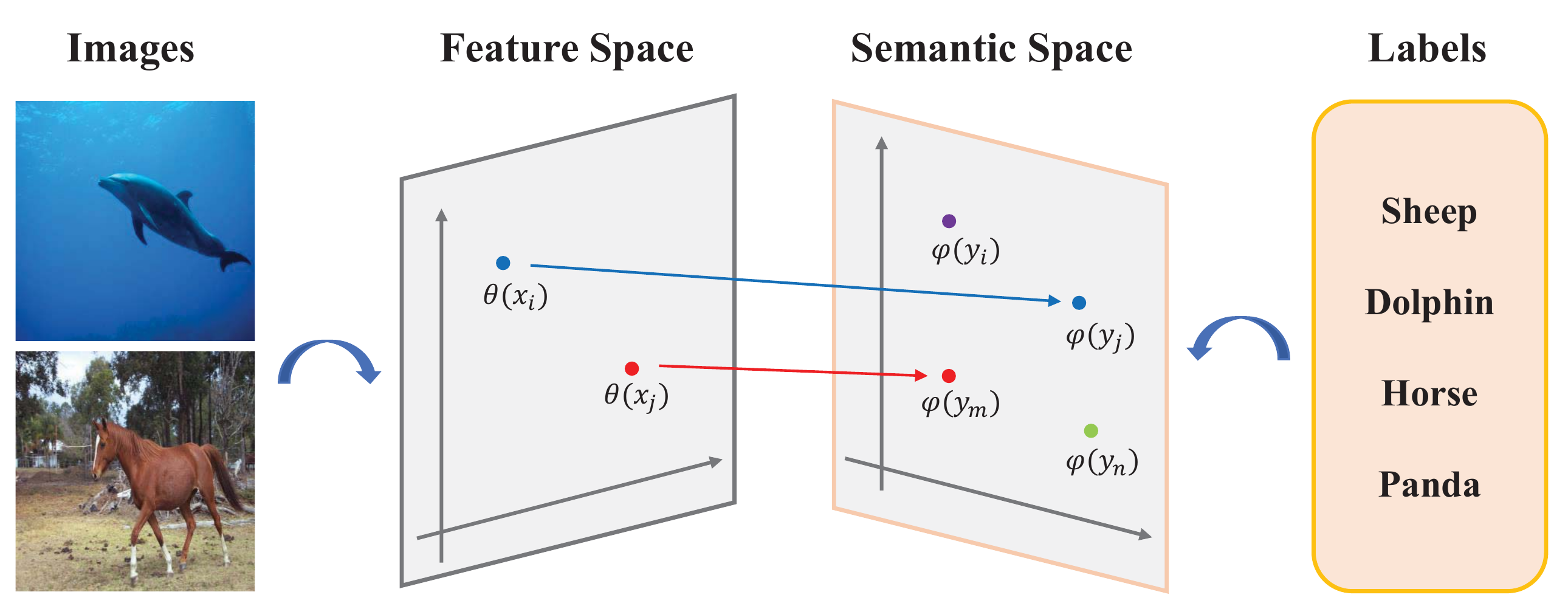}
	}
	\subfigure[Probability-Prediction Strategy]{
		\label{fig_zslframe_2}
		\includegraphics[width = 0.43\textwidth]{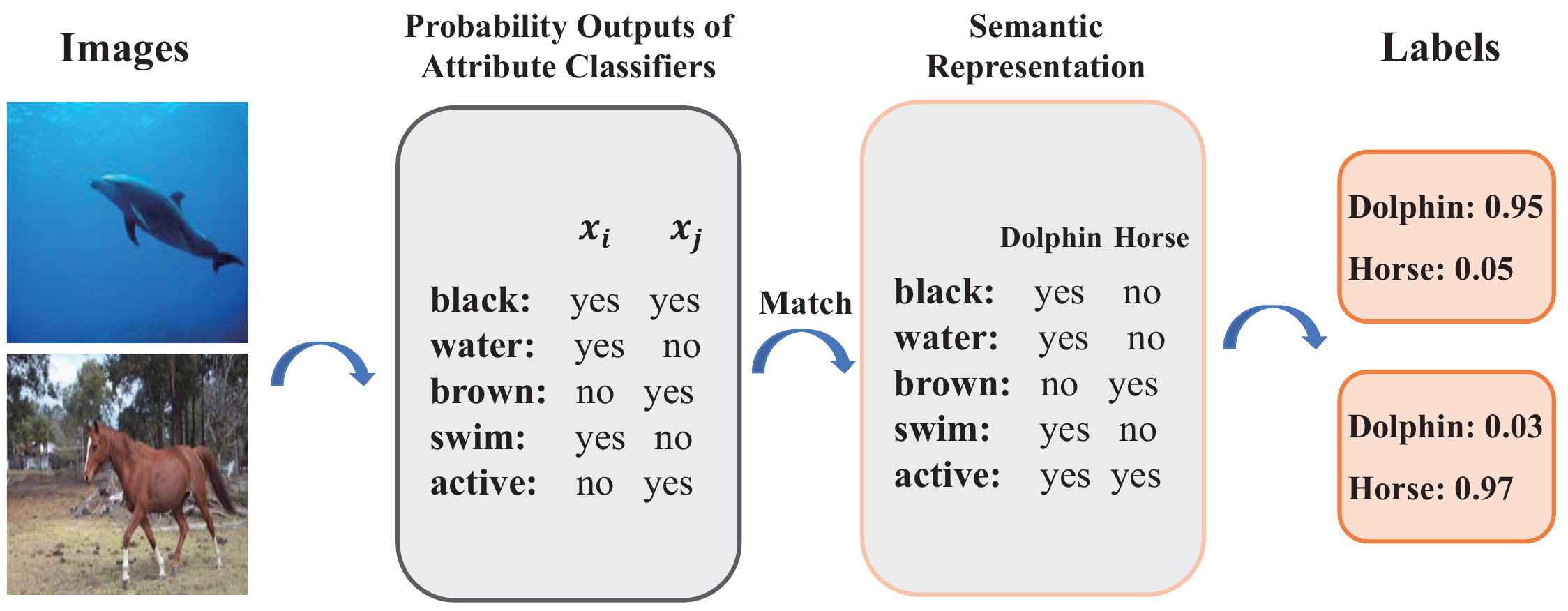}
	}
	\caption{Two kinds of ZSL models based on the label-embedding strategy and the probability-prediction strategy respectively.}
	\label{fig_zslframe}
\end{figure}

Object recognition is an attractive and practical research task in the computer vision field \cite{zhang2018transfer}. With the rapid development of machine learning technologies, especially the rise of deep neural networks, visual object recognition has made tremendous progress in recent years \cite{wei2017cross, sung2018learning}. However, these recognition systems require a massive amount of labeled training data to perform excellently, and there is a big challenge that these systems do not work properly when test data belongs to the different category of training data. On the contrary, given the description, humans can identify new objects even though they have not seen them in advance \cite{murphy2004big}. Taking the challenge into account and inspired from the human cognition system, zero-shot learning (ZSL) has received increasing attention in recent years \cite{liu2018deep, qi2017joint}. 

Zero-shot learning aims to recognize new unseen objects by using disjoint seen objects \cite{lampert2014attribute}, thus it relies on some high-level semantic representation of objects to share information between different categories. As a kind of widely used semantic representation, attributes are typical nameable properties of objects \cite{farhadi2009describing}. Attributes could be the color or the shape of objects, or they could be a certain part or a manual description of objects. For example, object \textit{elephant} has attributes \textit{big} and \textit{long nose}, object \textit{zebra} has the attribute \textit{striped}. Attributes can be binary values for describing the presence or absence of properties for objects, or be continuous values for indicating the correlation between properties and objects. 

Attributes play the key role in sharing information between the seen and unseen classes, and consequently govern the performance of ZSL model \cite{wang2016category}. In most of the recent ZSL methods, all the attributes are utilized and treated equally to recognize objects, whether they are relevant or not \cite{lampert2014attribute}. However, in fact, some attributes may be irrelevant to some specific objects since attributes are designed to describe the entire dataset \cite{guo2018zero}. For example, attribute \textit{small} has great correlation with object \textit{rat}, while it has little relatedness with object \textit{elephant}. Intuitively, relevant attributes will have greater weights than irrelevant attributes in object recognition systems. Therefore, relevant attributes will contribute a large amount of information for transferring information, while the irrelevant attributes would contribute little\cite{fu2012attribute}. For example, attribute \textit{water} can play a critical role in recognizing object \textit{fish}, while it is nearly useless for recognizing object \textit{bird}.

\begin{figure}[t] 
	\centering
	\includegraphics[width = 0.43\textwidth]{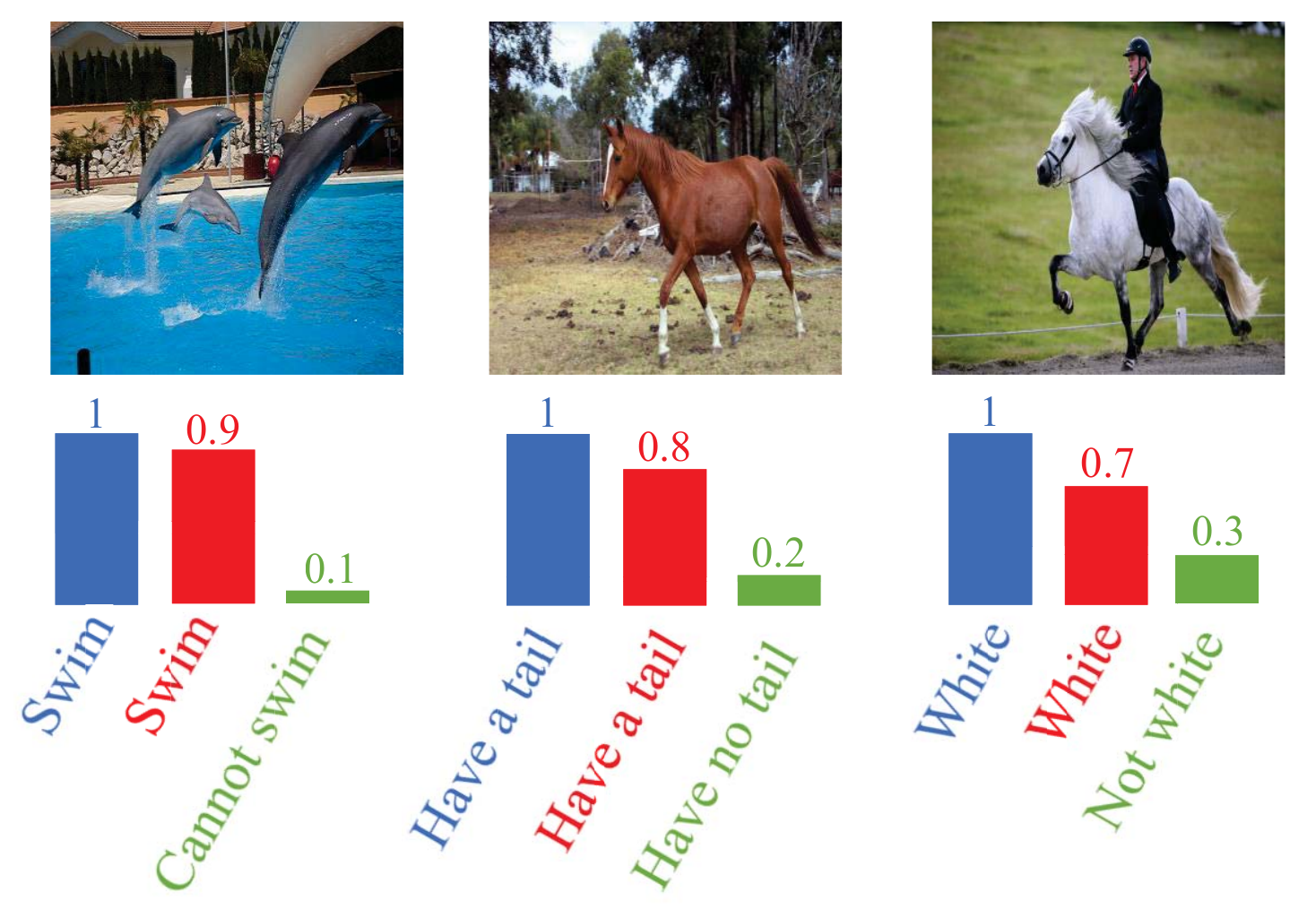}  
	\caption{Original binary value attributes (blue), normalized continuous value attributes (red) and complementary attributes (green) derived from AwA2 dataset. Normalized attributes and complementary attributes are used in this work.}
	\label{fig_cadef}
\end{figure}

In order to make all the attributes useful and efficient for zero-shot learning, in this paper, we introduce a notion of complementary attributes (CA) as the supplement to the original attributes. Complementary attributes are the opposite form of the original attributes as shown in Fig. \ref{fig_cadef}. For example, the complementary attribute corresponding to the attribute \textit{small} is \textit{not small}, and corresponding to the attribute \textit{have a tail} is \textit{have no tail}. In this way, complementary attributes will be more appropriate to describe objects when the corresponding original attributes are irrelevant, and thus CA can take full advantage of the semantic information contained in attributes. In other words, complementary attributes and original attributes are complementary to describe objects more comprehensively, and consequently make the ZSL model more efficient and robust.

Since the complementary attributes are the supplement to the original attributes, we can easily improve existing attribute-based ZSL methods by expanding original attributes with complementary attributes. Recent ZSL methods are usually based on two kinds of strategies, i.e. the label-embedding strategy based ZSL model (LEZSL) \cite{akata2015evaluation} and the probability-prediction strategy based ZSL model (PPZSL) \cite{lampert2014attribute}, as shown in Fig. \ref{fig_zslframe}. In LEZSL, recognition system learns a mapping function from the image feature space to the attribute embedding space. In this case, we can directly expand the original attribute representation by complementary attributes to improve the representation ability of the attribute embedding space. 

In PPZSL, ZSL system firstly pre-trains attribute classifiers based on seen data, and then use the nearest neighbor search mechanism to infer the label of unseen data. This system can effectively recognize zero-shot objects, while it suffers from a strong assumption that all the utilized attributes are independent of each other, which is impossible in reality \cite{akata2016label}. To solve this problem, we propose a novel rank aggregation (RA) framework to recognize unseen images. In the rank aggregation framework, the nearest neighbor search problem is transformed to the rank aggregation problem, which circumvents the assumption of attribute independence.

\begin{figure*}[htbp] 
	\centering
	\includegraphics[width = 0.9\textwidth]{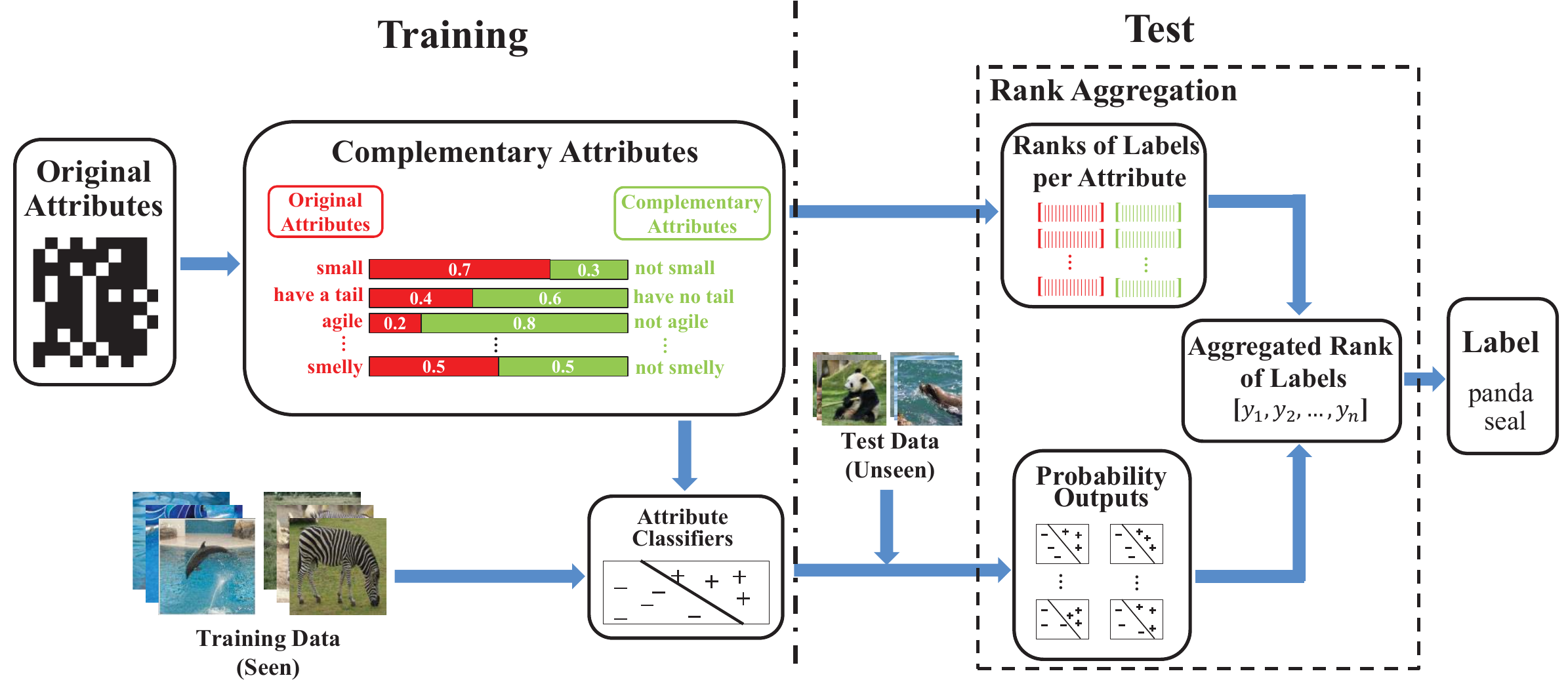}  
	\caption{The procedures of ZSL model using complementary attributes and rank aggregation framework. \textit{In training stage (left)}, classifiers of both original attributes and complementary attributes are trained using seen training data. \textit{In test stage (right)}, the label of test sample is predicted via rank aggregation framework which combines the ranks of labels per attribute with the probability outputs of attribute classifiers.}
	\label{fig_procedures}
\end{figure*}

In summary, we introduce complementary attributes as the supplement to the original attributes, and propose the rank aggregation framework to improve the efficiency and robustness of existing ZSL methods. The improved ZSL model via complementary attributes and rank aggregation is illustrated in Fig. \ref{fig_procedures}. To evaluate the effectiveness of the proposed model, extensive experiments are conducted on five ZSL benchmark datasets and the large-scale ImageNet dataset. Experimental results demonstrate that complementary attributes and rank aggregation can effectively and robustly improve current ZSL methods.

The main contributions of this work are summarized as follows:

1). To take full advantage of attributes, we introduce complementary attributes, as a supplement to the original attributes, to enhance the representation ability of semantic embedding space. 

2). We provide nontrivial theoretical analyses for complementary attributes. These analyses demonstrate that adopting CA is a principled way to improve the PAC-style generalization bound with theoretical guarantees.

3). For the probability-prediction strategy based ZSL model, there is an assumption that attributes are independent of each other. To solve this problem, we propose the rank aggregation framework to circumvent the assumption.

The rest of the paper is organized as follows. We review related works in Section \ref{sec_relatedworks}, introduce the complementary attributes in Section \ref{sec_method}, give the bound analysis in Section \ref{sec_bound}. The application of CA to existing ZSL models is presented in Section \ref{sec_modifymodel}. Experimental results and analyses are reported in Section \ref{sec_exper}. Section \ref{sec_conclu} draws conclusions.

\section{Related Works}
\label{sec_relatedworks}
Attributes are widely used as semantic representation for zero-shot object classification tasks. In this section, we review related works on attribute representation and zero-shot classification.

\subsection{Attribute Representation}
Attribute representation, as the visual properties of object description, is a kind of semantic information \cite{farhadi2009describing}. Attributes have been used as an intermediate representation to share information between different objects in various computer vision tasks, such as zero-shot
object classification \cite{lampert2014attribute,akata2016label}, action recognition \cite{cao2018body} and face recognition \cite{jang2018facial}.
In zero-shot learning, Lampert et al. \cite{lampert2014attribute} proposed the direct attribute prediction (DAP) model which categorizes zero-shot visual objects using attribute-label relationship as the assistant information. Akata et al. \cite{akata2016label} proposed the attribute label embedding (ALE) model which learns a compatibility function mapping image feature to attribute-based label embedding. Different from the leverage of definite value attribute, Parikh et al. \cite{parikh2011relative} presented relative attribute to describe objects. They proposed a model in which the supervisor relates the unseen objects to previously seen objects. Moreover, Wang et al. \cite{wang2016relative} proposed a relative attribute SVM for age estimation. Some other researchers utilized attributes for weakly supervised dictionary Learning \cite{wu2016exploiting}, partially labeled decision system \cite{dai2017attribute} and Face recognition \cite{jang2018facial}.

In existing attribute-based ZSL models, all the attributes are assumed to be effective and treated
equally. However, as mentioned in \cite{guo2018zero}, different attributes have different correlation with objects. Relevant attributes will be emphasized in transferring knowledge between different classes, while irrelevant attributes will be suppressed. Therefore, we introduce complementary attributes as the supplement to the original attributes to take full advantage of irrelevant attributes.

\subsection{Zero-shot Classification}
Zero-shot learning has the ability to recognize new objects by transferring knowledge from seen classes to unseen classes. Increasing attention has been attracted on ZSL and a lot of ZSL methods have been proposed in recent years. Based on the used strategy, most of recent ZSL methods can be divided into two kinds. 

One is the label-embedding strategy based ZSL methods. These models learn various functions that directly map input image features to semantic embedding space. Some researchers learned linear compatibility functions. Akata et al. \cite{akata2016label} presented an attribute label embedding model which learns a compatibility function using ranking loss. Frome et al. \cite{frome2013devise} presented a new deep visual-semantic embedding model using semantic information gleaned from unannotated text. Akata et al. \cite{akata2015evaluation} presented a structured joint embedding framework that recognizes images by finding the label with the highest joint compatibility score. Romera-Paredes et al. \cite{romera2015embarrassingly} proposed an approach which models the relationships among features, attributes and classes as a two linear layers network. Other researchers learned nonlinear compatibility functions. Xian et al. \cite{xian2016latent} presented a nonlinear embedding model that augments bilinear compatibility model by incorporating latent variables. For LEZSL, proposed complementary attributes can be easily applied to improve the model. We can directly concatenate the original attribute representation with CA to expand the attribute embedding space.

The other is the probability-prediction strategy based ZSL methods. These models adopt a two-steps framework and predict the label of test data by combining the probability outputs of pre-trained attribute classifiers. Lampert et al. \cite{lampert2014attribute} proposed two popular benchmark models, i.e. DAP and indirect attribute prediction (IAP). DAP learns probabilistic attribute classifiers using the seen data and infers the labels of unseen data by combining results of pre-trained classifiers. IAP induces a distribution over the labels of test data through the posterior distribution of the seen data by means on the class-attribute relationship. Similar to IAP, Norouzi et al. \cite{norouzi2014zero} proposed a method using convex combination of semantic embedding with no additional training. PPZSL can effectively recognize unseen objects by sharing attributes. However, it suffers from an assumption that all the attributes are independent of each other \cite{lampert2014attribute}. To solve this problem, we propose a novel rank aggregation framework to circumvents the assumption in this work.

\section{Complementary Attributes}
\label{sec_method}
In this section, we first formalize the ZSL task and then introduce the notion of the complementary attributes.

\subsection{ZSL Model Formulation}
\label{zslmodelformulation}
We consider zero-shot learning to be a task that classifies images from unseen classes by given images from seen classes. Given a training dataset $D_s = \{(x_n,y_n),n=1,...,N_{s}\}$ of input/output pairs with $x_n\in \mathcal{X}$ and $y_n\in \mathcal{Y}_s$, our goal is to learn a nontrivial classifier $f:\mathcal{X}\rightarrow \mathcal{Y}_u$, which maps test images to unseen classes. Since training classes $\mathcal{Y}_s$ (seen classes) and test classes $\mathcal{Y}_u$ (unseen classes) are disjoint, i.e. $\mathcal{Y}_s\cap \mathcal{Y}_u=\emptyset$, knowledge needs to be transferred between $\mathcal{Y}_s$ and $\mathcal{Y}_u$ via sharing attributes.

Given training data $D_s$ and attribute representation $\mathbf{A}$, we can learn the mapping function $f$ by minimizing the following regularized empirical risk:
\begin{equation}
L\left ( y,f\left ( x;\mathbf{W} \right ) \right ) = \frac{1}{N_{s}}\sum_{n=1}^{N_{s}}l\left ( y_{n},f\left ( x_{n};\mathbf{W} \right ) \right )+\Omega \left ( \mathbf{W} \right ),
\label{eq_lossofzsl}
\end{equation}
where $l\left ( . \right )$ is the loss function, which can be square loss $1/2(f(x)-y)^2$, logistic loss $log(1+exp(-yf(x)))$ or hinge loss $max(0,1-y(f(x)))$. $\mathbf{W}$ is the parameter of mapping $f$ and $\Omega\left ( . \right )$ is the regularization term.

To solve the optimization problem in \eqref{eq_lossofzsl}, we need to define the mapping function $ f $ from input image feature space to attribute embedding space. There are two popular strategies to handle the zero-shot learning task, and correspondingly $ f $ can be defined in two forms. 

On the one hand, some researchers adopt the probability-prediction strategy for ZSL. They learn probabilistic attribute classifiers as a supervised learning problem at training time, and make a class prediction by combining scores of the learned attribute classifiers at test time. The mapping function $f$ in this strategy is defined as follows:
\begin{equation}
f(x)=\arg\mathop{\max}_{y \in \mathcal{Y}_{u}}\prod _{m=1}^{M}\frac{p(a_m^y|x)}{p(a_m^y)},
\label{eq_foftwosteps}
\end{equation}
where $M$ is the number of attributes, $ a_m^y $ is the $ m^{th} $ attribute for class $ y $, and $\mathcal{Y}_{u}$ is the label of unseen classes. $p(a_m^y|x)$ is the posterior probability, and $p(a_m^y)$ is the prior probability.

On the other hand, most of resent researchers use the label-embedding strategy for ZSL. They directly learn the mapping function $f$ by optimizing the loss function in \eqref{eq_lossofzsl}. In this strategy, $f$ is defined as follows:
\begin{equation}
f\left ( x;\mathbf{W} \right )=\arg\mathop{\max}_{y\in \mathcal{Y} } F\left ( x,y;\mathbf{W} \right ),
\label{eq_foflabelembed}
\end{equation}
where function $F:\mathcal{X} \times \mathcal{Y}\rightarrow \mathbb{R}$ is a bi-linear compatibility function that associates images features and attribute representation, and can be defined as follows:
\begin{equation}
F\left ( x,y;\mathbf{W} \right )=\theta \left ( x \right )^{T}\mathbf{W} \varphi \left ( y \right ) ,
\label{eq_Flesori}
\end{equation}
where  $\theta \left ( x \right )$ is the image features of sample $x$, $ \varphi \left ( y \right ) $ is the label embedding (i.e. attribute representation) of class $y$.

We summarize some frequently used notations in Table \ref{table_notations}.

\begin{table}[t]
	\centering
	\renewcommand\arraystretch{1.2}
	\caption{Notations and Descriptions.}
	\label{table_notations}
	\setlength{\tabcolsep}{0.5mm}{
		\begin{tabular}{|c|c||c|c|}
			\hline
			\textbf{Notation} &\textbf{Description} &\textbf{Notation} &\textbf{Description} \\ \hline\hline
			$D_s$ &training dataset &$N_s$ &\#training samples \\ \hline
			$\mathcal{X}$ &image features &$d$    &\#dimension of features \\ \hline
			$\mathcal{Y}_s$ &training classes (seen)    &$K$ &\#training classes \\ \hline
			$\mathcal{Y}_u$ &test classes (unseen)       &$L$    &\#test classes \\ \hline
			
			$\mathbf{A}$ &original attribute matrix  &${M}$      &\#original attributes     \\ \hline
			$\overline{\mathbf{A}}$ &complementary attribute matrix    &${N_a}$ &\#all the attributes    \\ \hline 	
			$\mathbf{S}$ &expanded attribute matrix  & $\mathbf{W}$ &learned parameters  \\  \hline
			$\mathbf{a}_j$ &attribute representation of class $j$  &$\mathbf{r}$ &rank of labels  \\ \hline
	\end{tabular}}
\end{table}

\begin{figure*}[t]
	\centering 
	\subfigure[AwA]{ 
		\label{fig_entropy_1} 
		\includegraphics[width=0.18\textwidth]{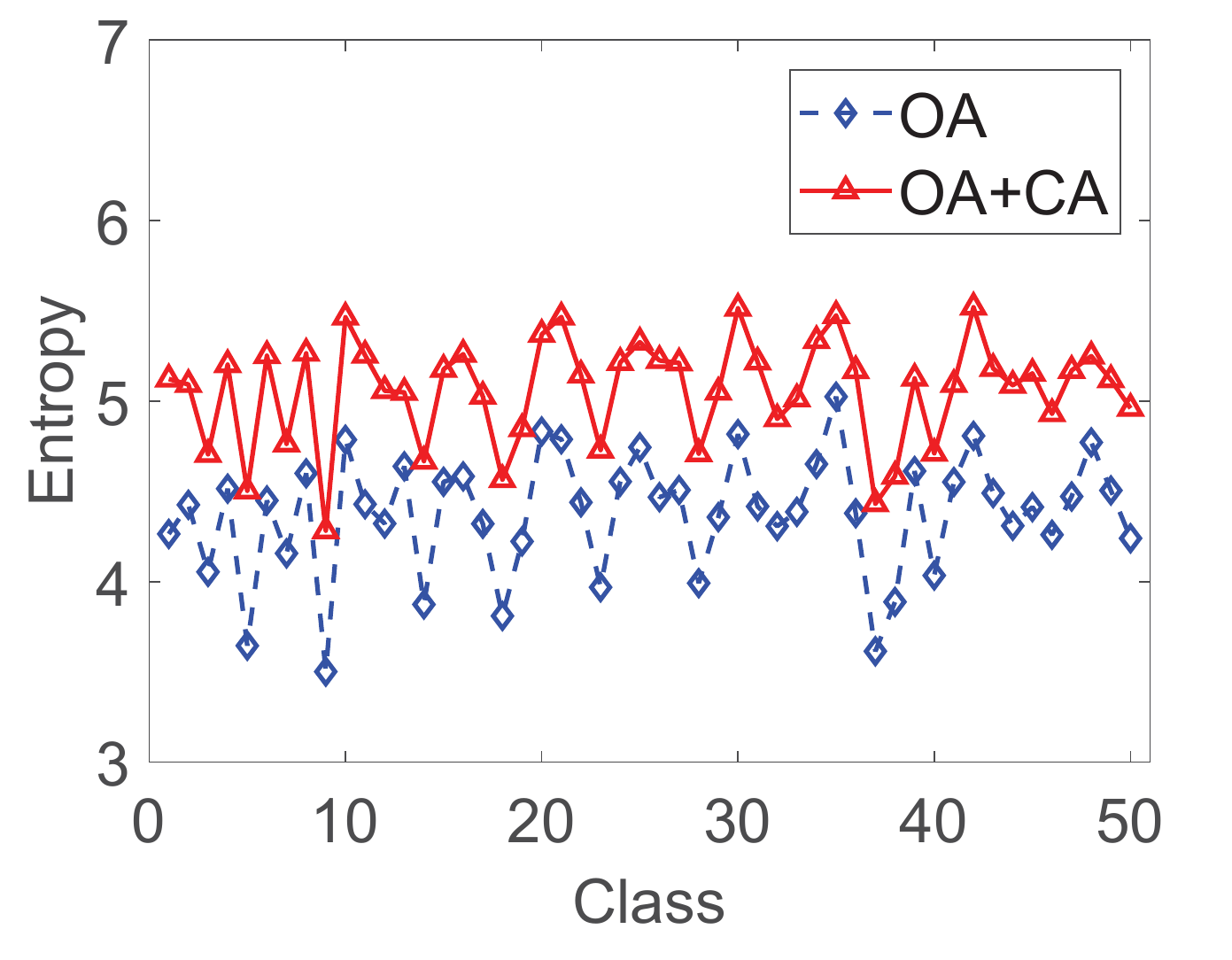} 
	} 
	\subfigure[AwA2]{ 
		\label{fig_entropy_2} 
		\includegraphics[width=0.18\textwidth]{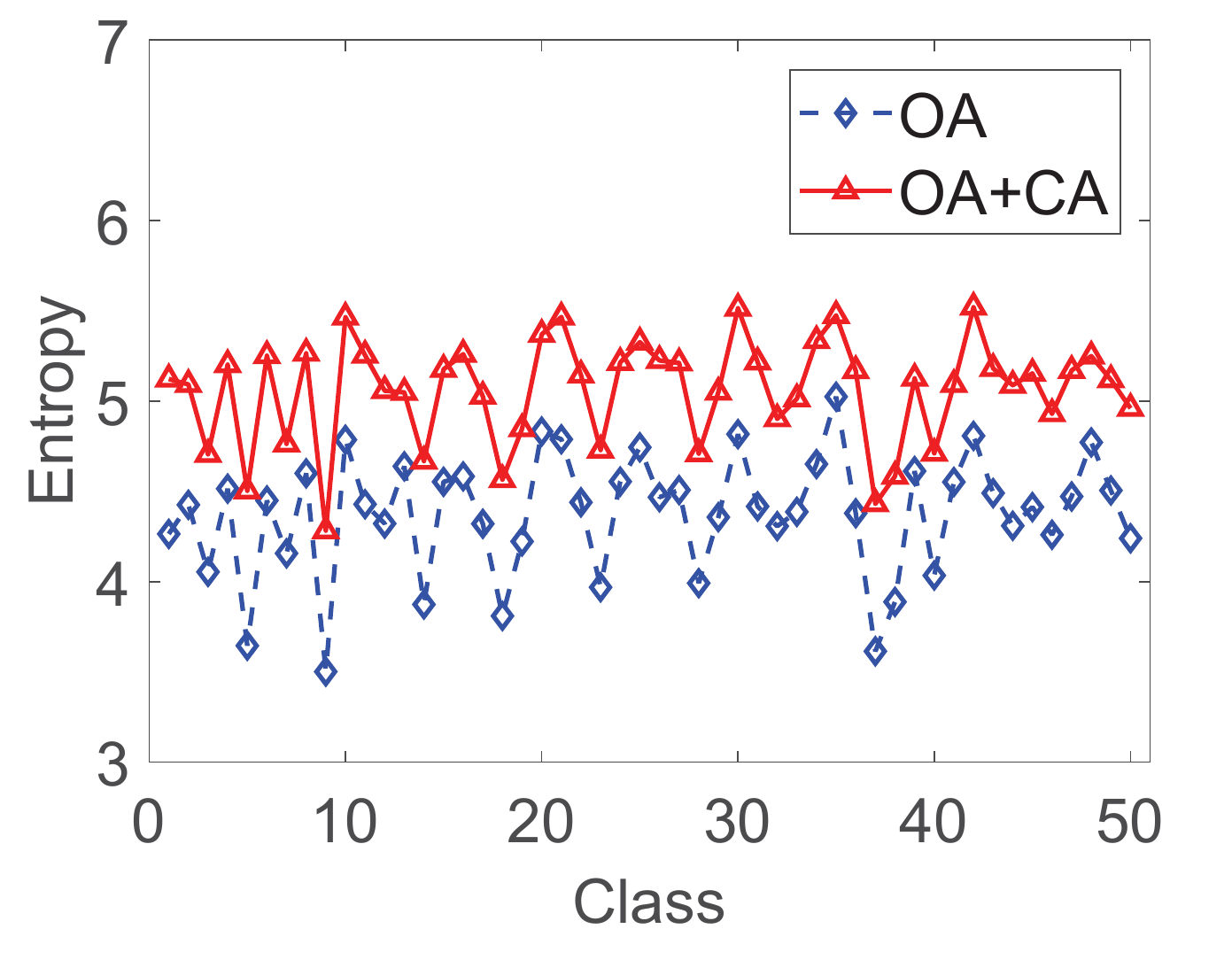} 
	} 
	\subfigure[aPY]{ 
		\label{fig_entropy_3} 
		\includegraphics[width=0.18\textwidth]{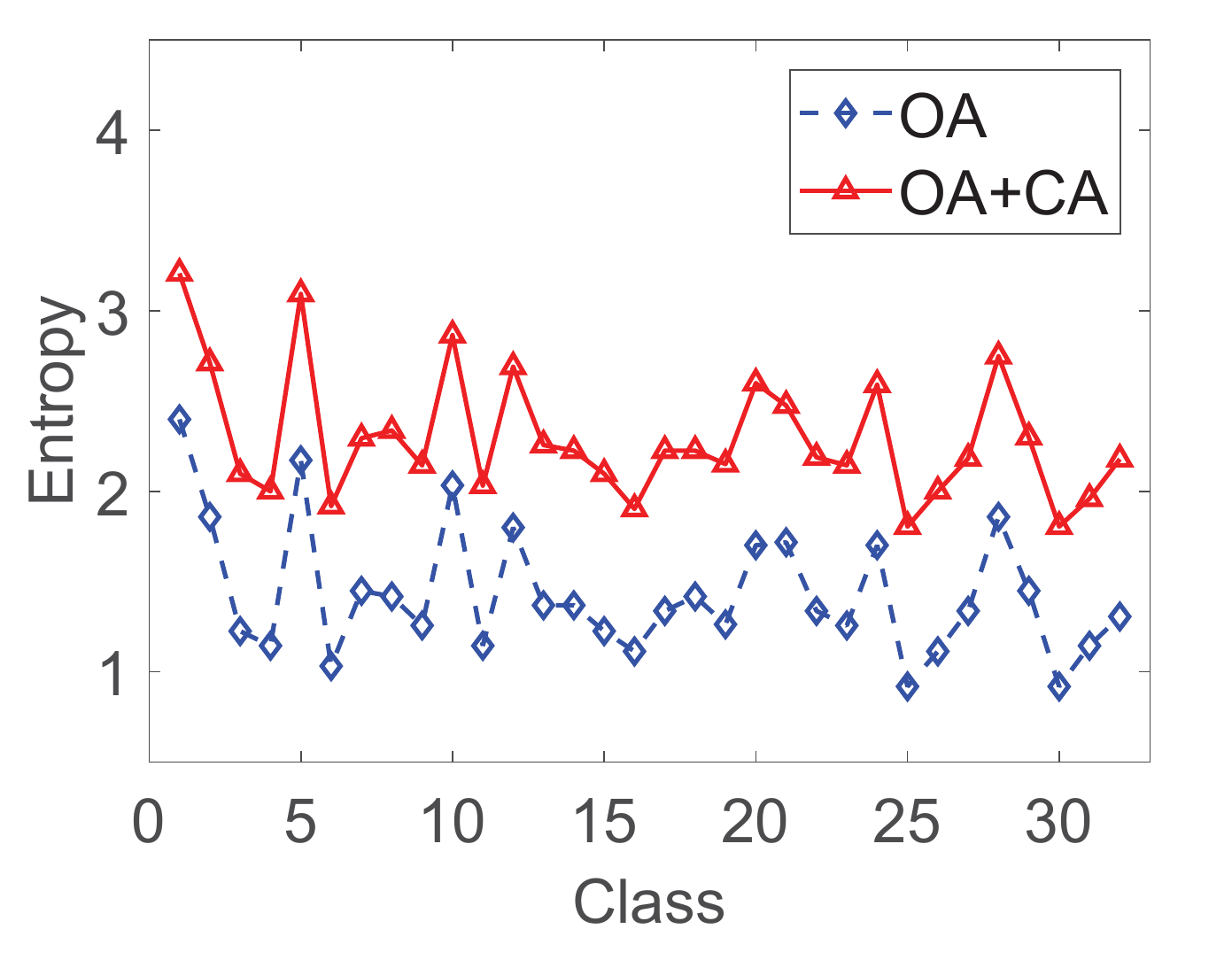} 
	} 
	\subfigure[CUB]{ 
		\label{fig_entropy_4} 
		\includegraphics[width=0.18\textwidth]{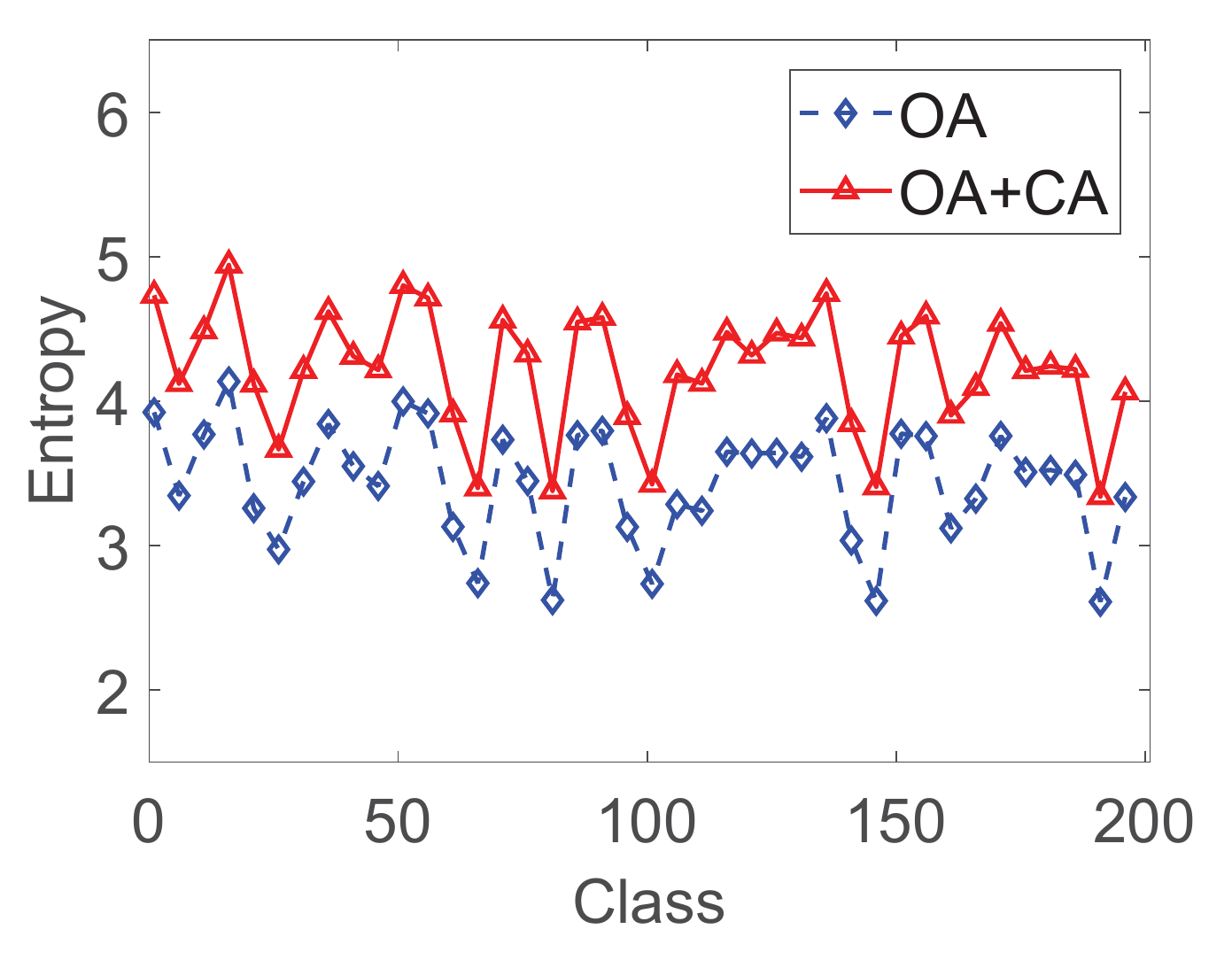} 
	} 
	\subfigure[SUN]{ 
		\label{fig_entropy_5} 
		\includegraphics[width=0.18\textwidth]{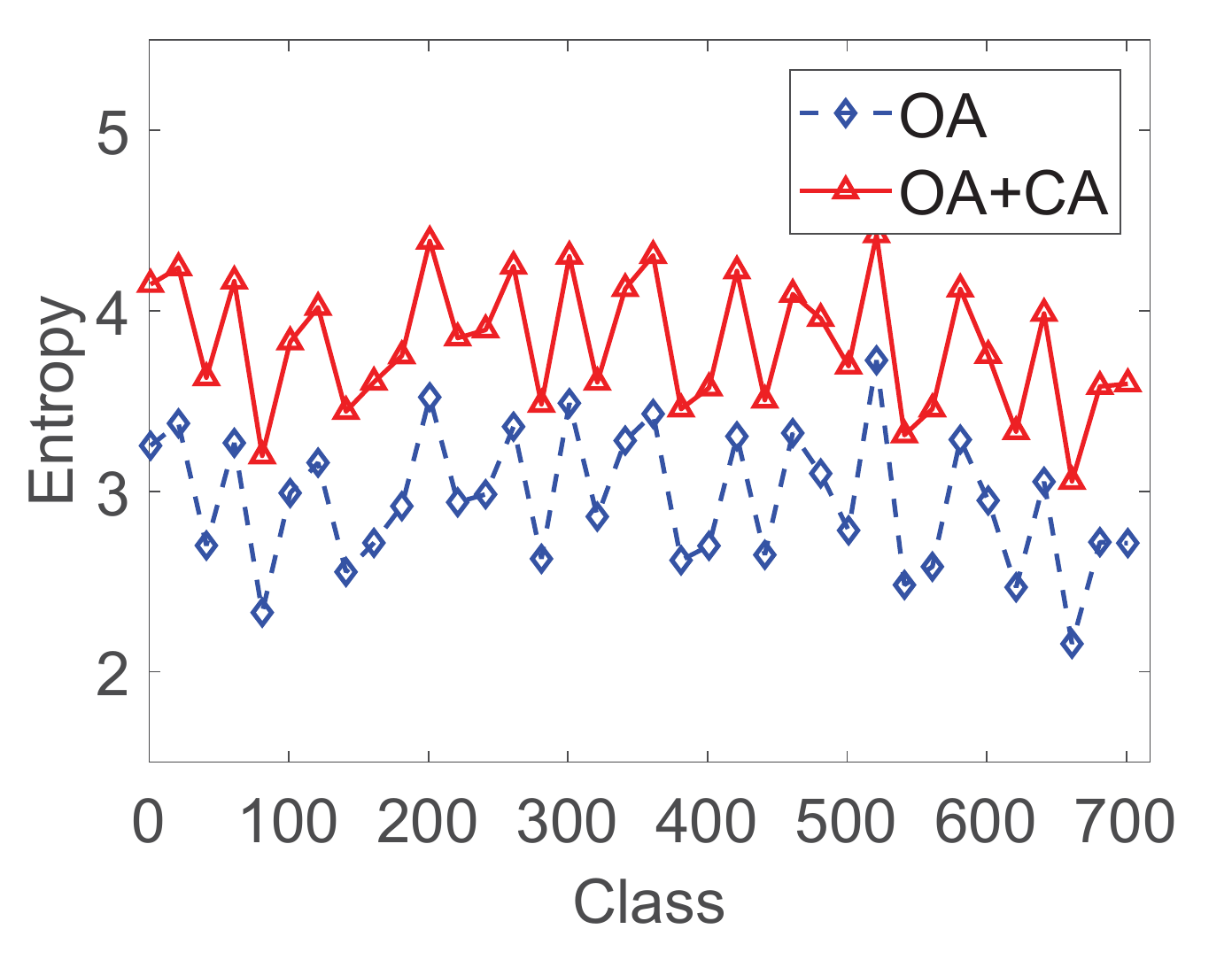} 
	} 	
	\caption{The information amount (measured by entropy) of original attributes (OA) and complementary attributes (CA) for each class on five datasets.}
	\label{fig_entropy} 
\end{figure*}

\subsection{Complementary Attributes}
Attributes, as semantic properties of visual objects, are used to solve zero-shot learning problem in many works \cite{farhadi2009describing,lampert2014attribute,akata2016label}. For example, $ a_m^y $ in \eqref{eq_foftwosteps} and $ \varphi \left ( y \right ) $ in \eqref{eq_Flesori} are different forms of the attribute representation that used in attribute-based ZSL methods. The value of the attribute indicates the presence or absence of semantic property for a specific object. For example, the attribute \textit{wing} is present with respect to the object \textit{birds} and is absent with respect to the object \textit{fish}. Attributes of ZSL datasets are designed to describe objects as clearly as possible. While parts of attributes have little correlation with some objects although these attributes may be more suitable for describing other objects. For example, attribute \textit{small} is effective for recognizing the object \textit{rat} while it has little correlation with the object \textit{elephant}. Therefore, in this work, we expand attributes with complementary attributes, as the supplement to the original attributes, to describe objects more comprehensively. For example, we add \textit{not small} as the complementary attribute for attribute \textit{small}, and obviously, the complementary attribute \textit{not small} is more relevant than the original attribute \textit{small} with respect to the object \textit{elephant}. Hence, complementary attribute (like \textit{not small}) can contribute more information for recognizing zero-shot object (like \textit{elephant}) when the original attribute (like \textit{small}) contribute little.

Suppose we have $K$ seen classes for training and $L$ unseen classes for testing. Each class has an attribute representation $\mathbf{a}_j=[a_j^{(1)},...,a_j^{(i)},...,a_j^{(m)}]^T$, where $a_j^{(i)}$ denotes the value of the $i^{th}$ attribute for class $j$. Original attribute matrix $\mathbf{A}$ consists of attribute vectors of all classes including training classes $\mathcal{Y}_s$ and test classes $\mathcal{Y}_u$ as follows:
\begin{equation}
\mathbf{A}=[{\mathbf{a}_\mathcal{Y}}_s,{\mathbf{a}_\mathcal{Y}}_u]=[\mathbf{a}_{\mathcal{Y}_{s1}},...,\mathbf{a}_{\mathcal{Y}_{sK}},\mathbf{a}_{\mathcal{Y}_{u1}},...,\mathbf{a}_{\mathcal{Y}_{uL}}],
\label{eq_A}
\end{equation}
where each column of matrix $\mathbf{A}$ represents a class. Then we can do l2-normalization to each column as follows \cite{xian2018zero}:
\begin{equation}
\mathbf{a}_j=\mathbf{a}_j/ \| \mathbf{a}_j  \|_2,
\label{eq:simil}
\end{equation}
and $a_{ij}$ in matrix $\mathbf{A}$ can be considered as the correlation between the $i^{th}$ attribute and class $j$.

After normalizing the original attribute matrix $\mathbf{A}$,  we can easily get the complementary attribute matrix $ \mathbf{\overline {A}} $ as follows:
\begin{equation}
\mathbf{\overline {A}}=\mathbf{1}-\mathbf{A},
\label{equation_s1ms}
\end{equation}  
where $\mathbf{1}$ is the matrix of ones (with the same shape as $\mathbf{A}$).

In order to leverage CA to enhance the semantic representation, we expand the original attribute matrix $ \mathbf{A} $ by the complementary attribute matrix $ \mathbf{\overline{A}} $.
The expanded attribute matrix $\mathbf{S}$ that concatenates $ \mathbf{A} $ and  $ \mathbf{\overline{A}} $ is calculated as follows:
\begin{equation}
\mathbf{S}=\begin{bmatrix}\mathbf{A}\\ \mathbf{\overline{A}}\end{bmatrix}.
\label{eq:expand}
\end{equation}

To evaluate the extra information gained from complementary attributes, we use the entropy to measure the information amount of attributes \cite{guo2018zero}. Entropy, as a statistical measure of randomness, is defined for each class as follows:
\begin{equation}
	En(\mathbf{a}_j)= \sum_{i}^M p_i \textrm{log} p_i,
	\label{eq_entropy}
\end{equation}
where $ p_i $ denotes the distribution of attribute $\mathbf{a}_j$ for class $j$. 
We plot the information amount of original attributes and complementary attributes for each class on five ZSL benchmark datasets in Fig. \ref{fig_entropy}. Obviously, adopting complementary attributes can increase the entropy, i.e. obtain more information, which suggests that the complementary attributes can be more efficient for sharing knowledge between different classes in ZSL. Therefore, complementary attributes can enhance the representation ability of semantic attribute space and make the attribute representation more discriminative.

The complementary attributes can be easily adopted to improve existing attribute-based ZSL methods. The modification of existing ZSL methods by complementary attribute is presented in Section \ref{sec_modifymodel}.

\section{Bound Analysis}
\label{sec_bound}
In this section, we analyze the PAC-style bound of proposed model which determines its ability to recognize samples from unseen classes. We first introduce the theoretical bound of original attribute-based ZSL model, and then analyze the bound changes after adopting complementary attributes.

\subsection{Bound for ZSL model}
To simplify the analysis, we consider that ZSL model adopts the probability-prediction strategy, and address the question that how many training samples are necessary to correctly recognize unseen samples with a high probability. Since ZSL model can be split into two steps, i.e.  predicting attributes and recognizing labels, we can address this question through two steps \cite{palatucci2009zero}. Firstly, we analyze how much error can be tolerated in predicting attributes when the ZSL model can still recognize unseen samples with a high probability. Then, we analyze how many training samples are needed to achieve the bound of the first step. Intuitively, if we can correctly predict the attributes for an unseen sample in the first step, then we will have a high probability of recognizing the unseen sample in the second step.

For simplicity, we assume that attributes are binary values, and the label recognition stage is a nearest-neighbor classifier with the hamming distance metric. We first analyze the amount of error we can tolerate when predicting attributes on unseen samples. For the unseen sample $x$, we can map it from the image feature space to the attribute representation space at the point p. Let $d(p,p')$ be the distance between the mapped point $p$ and another point $p'$ which represents a label embedding in the attribute space. Then we can define the relative distribution ${R}_p$ for the point $p$ as the probability that the distance $d(p,p')$ is less than some distance $z$ as follows:
\begin{equation}
	{R}_p(z)=P(d(p,p') \leq z).
	\label{eq_Rp}
\end{equation}

Suppose that $dn_p$ is the distance between the point $p$ and its nearest neighbor point, we can also define the relative distribution $G_p$ as follows:
\begin{equation}
{G}_p(z)=P(dn_p \leq z).
\label{eq_Gp}
\end{equation}

We can calculate $G_p$ using $R_p$ as shown in \cite{ciaccia2000pac}:
\begin{equation}
{G}_p(z)=1-(1-R_p(z))^n,
\label{eq_Gp2}
\end{equation}
where $n$ is the number of points in the attributes space, i.e. the number of unseen classes for ZSL task.

Here suppose that $dt_p$ is the distance between the mapped point $p$ and the point that represents the true label of sample $x$, we want a small probability $\gamma$ that the distance $dt_p$ is larger than $dn_p$:
\begin{equation}
P(dn_p \leq dt_p) \leq \gamma.
\label{eq_pr}
\end{equation}

Combining \eqref{eq_pr} with \eqref{eq_Gp}, we can get:
\begin{equation}
{G}_p(dt_p)\leq \gamma.
\label{eq_Gp3}
\end{equation}

The relation distribution $G_p(\cdot)$ is monotonically increasing since it is a cumulative distribution function. Hence, we can define a function  $G_p^{-1}(\gamma)$ as follows:
\begin{equation}
G_p^{-1}(\gamma) =\arg \mathop{\max}_{dt_p}G_p(dt_p)\leq \gamma.
\label{eq_Gpm1}
\end{equation}

Therefore, in the label recognition stage, if $dt_p \leq G_p^{-1}(\gamma)$, we can correctly recognize unseen samples with at least $(1-\gamma)$ probability.

After analyzing the bound of the label recognition stage, we need to analyze the bound of predicting attributes on unseen samples. In other words, how many training samples are needed to achieve the bound of $G_p^{-1}(\gamma)$.

Since we assume that the labels of unseen data are inferred by combining the outputs of $M$ attributes classifiers with a hamming distance metric, $G_p^{-1}(\gamma)$ indicates the number of incorrect prediction when predicting attributes on unseen samples. To simplify the analysis, we assume that each attribute classifier is PAC-learnable \cite{valiant1984theory} form an image feature space of $d$ dimensions. Then, we can define the tolerated error $\varepsilon $ for each attribute classifier as follows:
\begin{equation}
\varepsilon = G_p^{-1}(\gamma) / M,
\label{eq_errofeachcla}
\end{equation}
where $G_p^{-1}(\gamma)$ is the total error that can be tolerated, and $M$ is the number of attribute classifiers.

We analyze the PAC-style bound of each attribute classifier via the standard PAC bound for VC-dimension \cite{mitchell1997machine}. To obtain an attribute classifier with $(1-\delta)$ probability that has error rate at most $\varepsilon =  G_p^{-1}(\gamma) / M$, the classifier requires a number of training samples $N_{\delta}$ \cite{palatucci2009zero}:
\vskip -0.1in
\begin{small}
\begin{equation}
N_{\delta} \geq \frac{M}{G_p^{-1}(\gamma)}[4 \textrm{log}(2/\delta)+8(d+1)\textrm{log}(13M/G_p^{-1}(\gamma))],
\label{eq_Ndel}
\end{equation}
\end{small}
where $d$ is the dimension of image features.

Lastly, we analyze the bound of the overall ZSL model. The total probability \cite{palatucci2009zero} of correctly recognizing unseen samples is:
\begin{equation}
	P_{zsl} = (1-\delta)^M \cdot P_{att} \cdot (1-\gamma),
	\label{eq_pzsl}
\end{equation}
where
\begin{equation}
 P_{att} = 1-\textrm{BinoCDF}(M(1-\varepsilon),M,(1-\varepsilon)).
	\label{eq_patt}
\end{equation}

In \eqref{eq_pzsl}, $(1-\delta)^M$ is the probability that all the attribute classifiers will achieve the tolerated error rate $\varepsilon$, and $(1-\gamma)$ is the probability of correctly inferring the labels given the outputs of attribute classifiers. $ P_{att} $ is the probability that these $M$ attribute classifiers will make at least $M(1-\varepsilon)$ correct prediction. Specifically, 
 $\textrm{BinoCDF}(M(1-\varepsilon),M,(1-\varepsilon))$ is the Binomial Cumulative Distribution function\footnote{https://www.mathworks.com/help/stats/binocdf.html} \cite{wadsworth1960introduction}, where the attribute classifiers follow the Binomial Distribution with parameters $ M $ and $ (1-\varepsilon) \in [0,1]$, i.e. $ \textrm{B}(M, 1-\varepsilon) $.

In summary, given desired error parameters $\delta$ and $\gamma$ for two steps of a ZSL model, we can guarantee that this ZSL model can correctly recognize unseen samples with a probability of $P_{zsl}$ (in \eqref{eq_pzsl}) if provided at least $N_{\delta}$ (in \eqref{eq_Ndel}) seen samples for training \cite{palatucci2009zero}.

\subsection{Bound for ZSLCA model}
In the ZSL with complementary attributes (ZSLCA) model, we introduce complementary attributes to expand the attribute representation space. The framework of ZSLCA is similar to ZSL model except the dimension of attribute space. Therefore, we can obtain the bound of ZSLCA by rewriting \eqref{eq_Ndel} and \eqref{eq_pzsl} as follows:
\vskip -0.1in
\begin{small}
	\begin{equation}
	\hat{N}_{\delta} \geq \frac{2M}{2G_p^{-1}(\gamma)}[4 \textrm{log}(2/\delta)+8(d+1)\textrm{log}(13\cdot 2M/2G_p^{-1}(\gamma))],
	\label{eq_Ndelca}
	\end{equation}
\end{small}
\begin{equation}
\hat{P}_{zsl} = (1-\delta)^{M} \cdot \hat{P}_{att} \cdot (1-\gamma).
\label{eq_pzslca}
\end{equation}
where
\begin{equation}
\hat{P}_{att} = 1-\textrm{BinoCDF}(2M(1-\varepsilon),2M,(1-\varepsilon)).
\label{eq_pattca}
\end{equation}

We can get that $ \hat{P}_{att} > P_{att} $, since $\textrm{BinoCDF}(2M(1-\varepsilon),2M,(1-\varepsilon)) < \textrm{BinoCDF}(M(1-\varepsilon), M, (1-\varepsilon))$ \cite{wadsworth1960introduction}. Comparing \eqref{eq_Ndel} with \eqref{eq_Ndelca}, \eqref{eq_pzsl} with \eqref{eq_pzslca}, we can obtain that:
\begin{equation}
\hat{N}_{\delta} = {N}_{\delta},
\label{eq_Ndelcomp}
\end{equation}
\begin{equation}
\hat{P}_{zsl} > {P}_{zsl}.
\label{eq_pzslcomp}
\end{equation}

Therefore, after using complementary attributes, the attribute-based ZSL model can recognize unseen samples with a higher probability ($\hat{P}_{zsl}$) when providing the same number ($\hat{N}_{\delta}$) of training samples.

\section{ZSL with Complementary Attributes}
\label{sec_modifymodel}
We can easily improve any attribute-based ZSL methods via complementary attributes. In this section, we present how to apply complementary attributes to existing ZSL methods. According to the strategies used in the ZSL methods, i.e. the probability-prediction strategy \cite{lampert2014attribute} and the label-embedding strategy \cite{akata2016label}, we present the improved models in Section \ref{sec_twosteps} and \ref{sec_labelembedding}, respectively. After that, we give the complexity analysis of the improved models.

\subsection{Complementary Attributes for PPZSL}
\label{sec_twosteps}
Some ZSL methods adopt the probability-prediction strategy, which infers the label of test samples by combining the outputs of pre-trained attribute classifiers. Since attributes are assumed to be independent of each other, these ZSL methods compare the similarity between the outputs of attribute classifiers and attribute representation of labels using the nearest neighbor search mechanism as in \eqref{eq_foftwosteps}. However, as showed in \cite{akata2016label}, the correlation between attributes is inescapable and cannot be ignored. To solve this problem and combining with complementary attributes, we propose a novel rank aggregation framework to predict the labels at test time for probability-prediction strategy based ZSL methods. In the rank aggregation framework, the nearest neighbor search problem is transformed to the rank aggregation problem which circumvents the assumption of attributes independence. 

Suppose we have $N_a$ ($N_a=2M$) attributes including $M$ original attributes and $M$ complementary attributes. Attribute classifiers are trained using seen training data at training time. At test time, each trained attribute classifier outputs a probability estimation $p(a_m \vert x)$ for the test sample $x$. 

After calculating attribute matrix $\mathbf{S}$, $N_a$ ranks (i.e. $\{\mathbf{r}^{(m)}, m = 1, ..., N_a\}$) among labels of test classes for each attribute can be induced. The rank $\mathbf{r}^{(m)}$ with regard to the $m^{th}$ attribute is defined as follows:
\begin{equation}
\text{class} \ i \ \text{ranked above class} \ j\Leftrightarrow r_i^{(m)}>r_j^{(m)},
\label{eq:rank}
\end{equation}
where $r_i^{(m)}$ and $r_j^{(m)}$ are equal to $s_{mi}$ and $s_{mj}$ in matrix $\mathbf{S}$, respectively.

Different attributes have distinct similarities that induce different ranks of labels. After obtaining $N_a$ ranks with regard to $N_a$ attributes, we need to recover the proximal ranks $ \mathbf{r} $ as the weighted aggregation of the $N_a$ ranks for each test sample. Here, we adopt a mixture of regression model to calculate the proximal rank.

Giving $N_a$ ranks $\{\mathbf{r}^{(m)}, m = 1, ..., N_a\}$, we can calculate the proximal rank $ \mathbf{r} $ as follows:
\begin{equation}
\mathop{\max}_{\mathbf{r}}\sum_{m=1}^{N_a}p(a_m \vert x)p(\mathbf{r} \vert \mathbf{r}^{(m)}),
\label{eq_minr}
\end{equation}
where $p(a_m \vert x)$ is the posterior probabilities calculated by pre-trained attribute classifiers, $ p(\mathbf{r} \vert \mathbf{r}^{(m)}) $ is the similarity between $ \mathbf{r} $ and $ \mathbf{r}^{(m)} $ and is defined as follows:
\begin{equation}
p(\mathbf{r} \vert \mathbf{r}^{(m)}) = \textrm{exp} ({\dfrac {- \| \mathbf{r}-\mathbf{r}^{(m)}\|_2^2} { \sigma}}),
\label{eq_srrm}
\end{equation}
where $ \sigma $ is a hyperparameter for normalization.

After getting the proximal rank $\mathbf{r}$ by solving the problem of \eqref{eq_minr} for each test sample, we can retrieve the highest order in $\mathbf{r}$ as the label of the test sample.

Combining complementary attributes with rank aggregation, the procedure of improved PPZSL is given in Algorithm \ref{alg:alg1}.

\begin{algorithm}[h]
	\caption{PPZSL improved by CA and RA}
	\begin{algorithmic}[1]
		\REQUIRE ~~\\
		Training samples and test samples; \\
		Attribute representation matrix $\mathbf{A}$. \\
		\ENSURE ~~\\
		Labels of test samples. \\
		\vspace{2mm}
		\STATE Calculate correlation between attributes and classes using Eq. \eqref{eq:simil};
		\STATE Expand original attribute matrix with complementary attributes using Eq. \eqref{eq:expand};
		\STATE Train attribute classifiers using training data;
		\STATE Get probability estimate for test sample by pre-trained attribute classifiers;
		\STATE Rank labels of test classes for each attribute based on the expanded attribute matrix (Eq. \eqref{eq:rank});
		\STATE Calculate the optimized rank using rank aggregation framework (Eq. \eqref{eq_minr});
		\STATE Retrieve the highest order as the label of test sample.
	\end{algorithmic} 
	\label{alg:alg1}	
\end{algorithm}

\begin{table*}[htbp]
	\centering
	\renewcommand\arraystretch{1.2}
	\caption{Statistics for Five ZSL Benchmark Datasets. In Standard Split (SS), Images are Split into Training Images (Seen) and Test Images (Unseen); in Proposed Split (PS), Images are Split into Training Images (Seen), Test Images (Seen) and Test Images (Unseen).}
	\label{table_datasetstat}
	\begin{tabular}{|c||c|ccc|c|cc|ccc|}
		\hline
		{\multirow{2}{*}{\textbf{Dataset}}} & {\multirow{2}{*}{\textbf{\#Attributes}}}  & \multicolumn{3}{c|}{\textbf{Classes}}  & \multirow{2}{*} {\textbf{\#Images}} & \multicolumn{2}{c|}{\textbf{Images (SS)}} & \multicolumn{3}{c|}{\textbf{Images (PS)}} \\ 
		& & {\textbf{\#All}} &{\textbf{\#Training}} & {\textbf{\#Test}} &  & {\textbf{\#Training}} &{\textbf{\#Test}}&  {\textbf{\#Training}} & {\textbf{\#Test(seen)}} &{\textbf{\#Test(unseen)}} \\ \hline\hline
		{AwA} & {85} & {50} & {40} & {10} & {30475} & {24295} & {6180} &19832 &4958 &5685 \\ 
		{AwA2} &{85} & {50} & {40} & {10} & {37322} & {30337} & {6985} &23527 &5882 &7913 \\ 
		{aPY} & {64} &{32} & {20} & {12} &  {15339} & {12695} & {2644} &5932 &1483 &7924 \\ 
		{CUB} & {312} & {200} &{150} & {50} &  {11788} & {8855} & {2933} &7057 &1764 &2967 \\
		{SUN} &102  &717 &645 &72 &14340   & 12900 &1440 &10320 &2580 &1440 \\
		\hline
	\end{tabular}	
\end{table*}

\begin{table*}[htbp]
	\centering
	\renewcommand\arraystretch{1.2}
	\caption{Conventional Zero-Shot Classification Accuracy (in \%) Using Standard Split (SS) and Proposed Split (PS). Numbers in Brackets are Relative Performance Gains. Boldface Indicates the Best. `-' Indicates that no Reported Results are Available.}
	\label{table_zsl}
	\setlength{\tabcolsep}{1.5mm}{
		\begin{tabular}{|c||cc|cc|cc|cc|cc|}
			\hline
			\multirow{2}{*}{\textbf{Method}} &\multicolumn{2}{c|}{\textbf{AwA}} &\multicolumn{2}{c|}{\textbf{AwA2}} & \multicolumn{2}{c|}{\textbf{aPY}} & \multicolumn{2}{c|}{\textbf{CUB}} & \multicolumn{2}{c|}{\textbf{SUN}} \\
			& \textbf{SS} & \textbf{PS} & \textbf{SS}  & \textbf{PS}   & \textbf{SS}     & \textbf{PS}   & \textbf{SS}          & \textbf{PS}         & \textbf{SS}         & \textbf{PS}         \\ \hline\hline
			ESZSL \cite{romera2015embarrassingly} &74.7 &58.2 &75.6 &58.6      &34.4 &38.3 & 55.1 &53.9 & 57.3          &  54.5         \\
			SJE \cite{akata2015evaluation} &76.7   &65.6     &69.5 &61.9       &  32.0          & 32.9          & 55.3           &  {53.9}         & 57.1          & 53.7          \\
			UVDS \cite{long2017zero} & 82.1          &-      &- &-    &  53.2         &-        &  45.7         &-        &-        &-        \\
			LESD \cite{ding2017low} &  82.8         &-     &- &-     &55.2       &  -    &45.2       & -        &    -       &   -  \\
			LESAE \cite{liu2018zero} &- &66.1 &- &68.4 &- &\textbf{40.8} &- &53.9 &- &60.0   \\
			SPAEN \cite{chen2018zero} &- &- &- &58.5 &- &24.1 &- &55.4 &- &59.2    \\ \hline
			DAP \cite{lampert2014attribute} &57.1          &44.1   &58.7 &46.1        &35.2         &33.8   &37.5     &40.0     &38.9       &39.9    \\
			DAP+CA & 69.7(+\textit{12.6}) &54.4(+\textit{10.3}) &70.2(+\textit{11.5}) &55.2(+\textit{9.1}) &39.6(+\textit{4.4}) &39.7(+\textit{5.9}) &43.5(+\textit{6.0}) &40.2(+\textit{0.2})  &41.3(+\textit{2.4}) &45.8(+\textit{5.9})   \\ \hline
			LatEm \cite{xian2016latent} &74.8 &55.1 &68.7 &55.8 &34.5 &35.2 &49.4 &49.3 &56.9 &55.3  \\
			LatEm+CA &77.8(+\textit{3.0}) &52.9(-\textit{2.2}) &74.1(+\textit{5.4}) &59.3(+\textit{4.5})  &38.8(+\textit{4.3}) &\textbf{40.8}(+\textit{5.6}) &49.6(+\textit{0.2}) &54.5(+\textit{5.2}) &58.5(+\textit{1.6}) &57.8(+\textit{2.5})  \\ \hline
			DEM \cite{zhang2017learning} &- &68.4 &- &67.1 &- &35.0 &- &51.7    &- &61.9        \\
			DEM+CA &- &69.7(+\textit{1.3}) &- &69.2(+\textit{2.1}) &- &38.5(+\textit{3.5}) &- &52.3(+\textit{0.6}) &- &53.8(+\textit{1.9}) \\ \hline
			MFMR \cite{xu2017matrix} &86.5 &68.2 &83.2 &65.0 &54.6 &30.6 &43.8 &39.1 &54.6 &57.8          \\
			MFMR+CA &\textbf{87.5}(+\textit{1.0}) &68.3(+\textit{0.1}) &\textbf{85.4}(+\textit{2.2}) &67.5(+\textit{2.5}) &\textbf{56.9}(+\textit{2.3}) &33.2(+\textit{2.6}) &44.2(+\textit{0.4}) &39.2(+\textit{0.1}) &55.0(+\textit{0.4}) &58.1(+\textit{0.3}) \\ \hline
			SEZSL \cite{kumar2018generalized} &83.8 &69.5 &80.8 &69.2 &- &- &60.3 &59.6 &64.5 &63.4          \\
			SEZSL+CA &85.3(+\textit{1.5}) &\textbf{70.9}(+\textit{1.4}) &81.8(+\textit{1.0}) &\textbf{70.4}(+\textit{1.2}) &- &- &\textbf{63.1}(+\textit{2.8}) &\textbf{62.2}(+\textit{2.6}) &\textbf{67.9}(+\textit{3.4}) &\textbf{64.1}(+\textit{0.7}) \\ \hline
	\end{tabular}}
\end{table*}

\subsection{Complementary Attributes for LEZSL}
\label{sec_labelembedding}
For the label-embedding strategy based ZSL model, which directly learns a mapping function from image feature space to label embedding space, complementary attributes can be easily applied to improve the performance.

At training time, the parameter $\mathbf{W}$ of the mapping function $f$ can be learned by minimizing the empirical risk defined in \eqref{eq_lossofzsl}. Specifically, the loss function $l$ for training sample $(x_n,y_n)$ is defined as follows:
\vskip -0.15in
\begin{small}
	\begin{equation}
	l = \sum_{y\in \mathcal{Y}_{s}}r_{ny} [ \triangle\left ( y_{n},y \right )+F\left ( x_{n},y;\mathbf{W} \right ) -F\left ( x_{n},y_{n};\mathbf{W} \right ) ]_{+},   
	\label{eq_LofLB}
	\end{equation}
\end{small}
where $\triangle(y_n;y)=1$ if $y_n = y$ or 0 otherwise. $r_{ny}\in [0,1]$ is the weight defined by different specific methods. For example, in LatEm \cite{xian2016latent}, $r_{ny}=1$, and in ALE \cite{akata2016label},  $r_{ny}$ is a ranking based weight.

After adopting complementary attributes, the learning algorithm and predicting function of the improved model is the same as the original model's. We just need to modify the label embedding $\varphi(y)$ in \eqref{eq_Flesori} by complementary attribute similarity matrix $\mathbf{S}$ as follows:
\begin{equation}
\varphi(y_j) = \mathbf{s}_j,
\end{equation}
where $\mathbf{s}_j$ is the $j^{th}$ column in  $\mathbf{S}$, i.e. the attribute representation of class $y_j$.

Using the modified label embedding $\varphi(y)$, we can learn the parameter $\mathbf{W}$ of mapping function $f$ by minimizing the loss function in \eqref{eq_LofLB}. At test time, we can predict the label of unseen samples via \eqref{eq_foflabelembed}.

\subsection{Complexity Analysis}
Suppose that there are $n$ test samples belonging to $L$ unseen classes, and the number of original attributes is $M$. The complexity of original ZSL model is $\mathcal{O}(nML^2)$, and it changes to  $\mathcal{O}(n \cdot 2M \cdot L^2) \sim \mathcal{O}(nML^2)$ after adopting complementary attributes. The complexity after adopting complementary attributes is the same as the original ZSL model's, which means that adopting complementary attributes does not increase the complexity of original ZSL model.

\section{Experimental Results}
\label{sec_exper}
To evaluate the efficacy of the proposed method, extensive experiments are conducted on five ZSL benchmark datasets (AwA, AwA2, aPY, CUB, SUN) and one large-scale dataset (ImageNet). In this section, we first compare the proposed method with several state-of-the-art baselines and report the results on conventional ZSL task and generalized ZSL task respectively. Then, we give the ablation analyses to further evaluate the proposed complementary attributes and rank aggregation.

\begin{figure}[t]
	\centering
	\subfigure[AwA2]{
		\label{figure_images_1}
		\includegraphics[width = 0.22\textwidth]{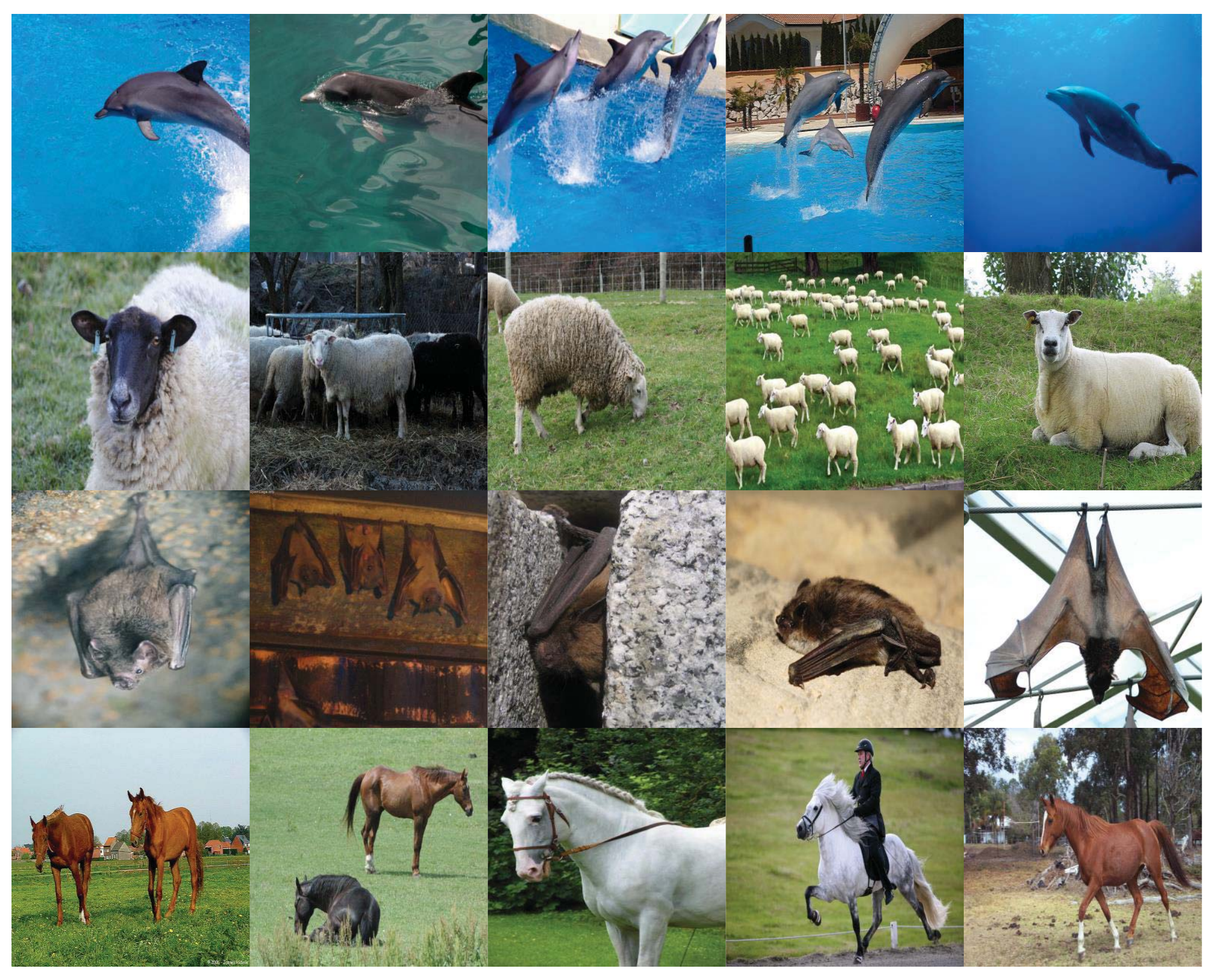}
	}
	\subfigure[aPY]{
		\label{figure_images_2}
		\includegraphics[width = 0.22\textwidth]{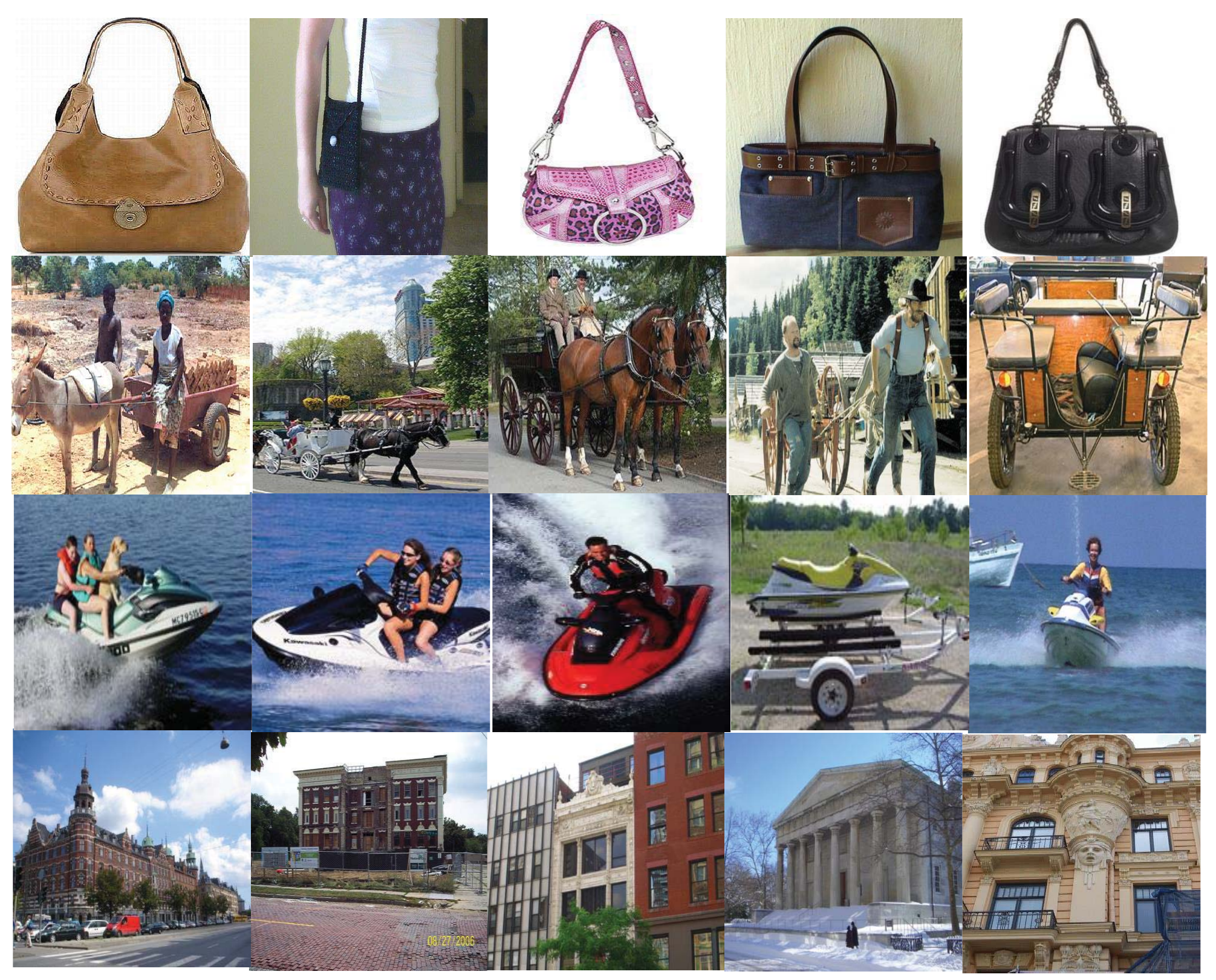}
	}
	\subfigure[CUB]{
		\label{figure_images_3}
		\includegraphics[width = 0.22\textwidth]{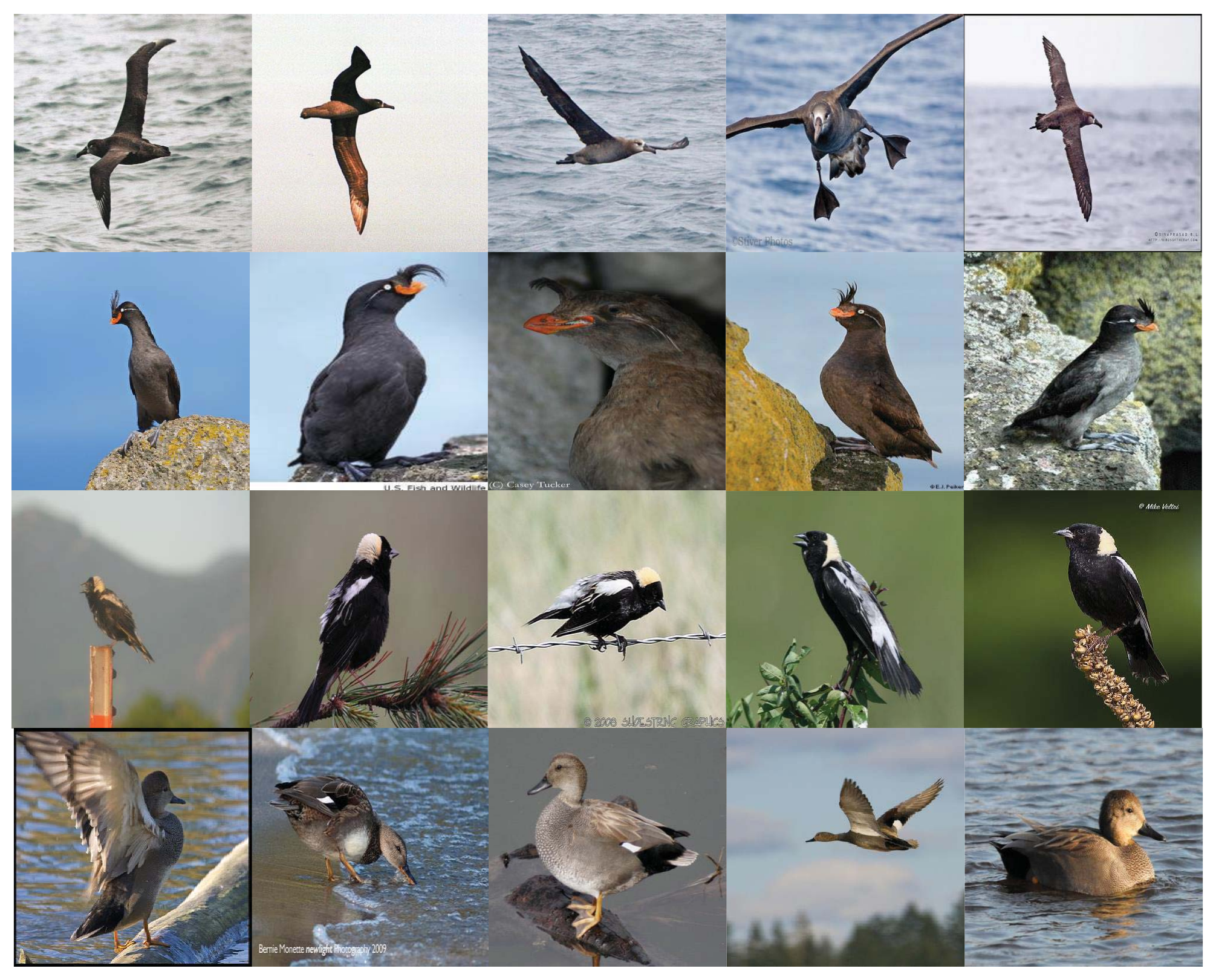}
	}
	\subfigure[SUN]{
		\label{figure_images_4}
		\includegraphics[width = 0.22\textwidth]{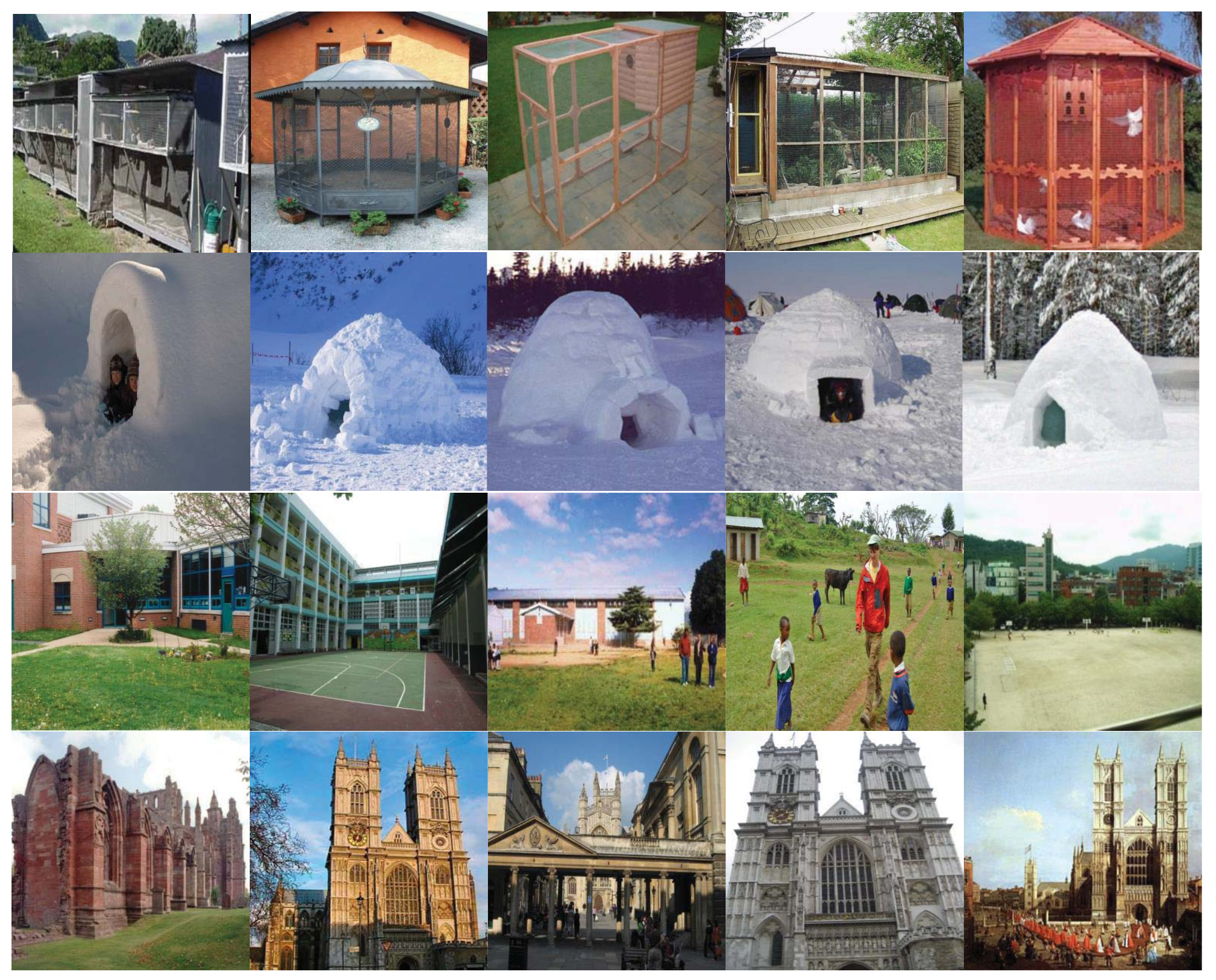}
	}
	\caption{Gallery of images from AwA2, aPY, CUB and SUN datasets.}
	\label{figure_images}
\end{figure}

\subsection{Experimental Setup}
\subsubsection{Datasets}
Experiments are conducted on five ZSL benchmark datasets: (1) Animal with Attribute (AwA)~\cite{lampert2014attribute}, (2) Animal with Attribute 2 (AwA2)~\cite{xian2018zero}, (3) attribute-Pascal-Yahoo (aPY)~\cite{farhadi2009describing}, (4) Caltech-UCSD Bird 200-2011 (CUB)~\cite{welinder2010caltech}, and (5) SUN Attribute Database (SUN) \cite{patterson2012sun}. ResNet-101~\cite{he2016deep} is used to extract deep features for experiments. We use 85, 85, 64, 312 and 102 dimensional continuous value attributes for AwA, AwA2, aPY, CUB and SUN datasets, respectively. The overall statistic information of these five datasets is summarized in Table \ref{table_datasetstat}. Fig. \ref{figure_images} shows the images derived from these datasets, where images of AwA dataset are not available for copyright reasons. 

We also evaluate the performance of the proposed method on the large-scale ImageNet dataset \cite{deng2009imagenet,russakovsky2010attribute} which contains a total of 14 million images from 21K classes. Following \cite{changpinyo2016synthesized}, we evaluate the proposed method on three scenarios: (1) 2-$hop$ contains 1,509 unseen classes that are within two tree hops of the seen 1K classes according to the ImageNet label hierarchy; (2) 3-$hop$ contains 7,678 unseen classes that are within three tree hops of the seen 1K classes; (3) all 20K contains 20,345 classes from the entire ImageNet dataset.

\subsubsection{Dataset Splits}
Zero-shot learning assumes that training classes and test classes are disjoint. Actually, ImageNet, the dataset exploited to extract image features via deep neural networks, may include any test classes. Therefore, Xian et al. \cite{xian2018zero} proposed a new dataset splits insuring that none of the test classes appear in the dataset used to training extractor model. In this work, we evaluate the proposed method using both splits, i.e. the original standard splits (SS) and the proposed splits (PS).

\subsubsection{Evaluation Protocols}
For the conventional zero-shot setting, we use the mean class accuracy, i.e. per-class averaged top-1 accuracy, as the criterion of assessment. Mean class accuracy is calculated as follows: 
\begin{equation}
	acc=\frac{1}{L} \sum_{y\in\mathcal{Y}_u}\frac{\# \mathrm{correct\; predictions\; in\;} y}{ \# \mathrm{samples\; in\;} y},
\end{equation}
where $L$ is number of test classes, $\mathcal{Y}_u$ is the set comprised of labels of test classes. 

For the generalized zero-shot setting, after calculating the mean class accuracy of the training and test classes, we can compute the harmonic mean accuracy \cite{xian2018zero} as follows:
\begin{equation}
	H=\frac{2*acc(ts)*acc(tr)}{acc(ts)+acc(tr)}, 
\end{equation}
where $acc(ts)$ and $acc(tr)$ represent the accuracy of images from
test classes and training classes, respectively.

\subsubsection{Compared Baselines}
We compare our model with several representative and state-of-the-art ZSL baselines: (1) probability-prediction strategy based methods: DAP \cite{lampert2014attribute}, LESD \cite{ding2017low}, SYNC \cite{changpinyo2016synthesized}; (2) label-embedding strategy based methods: ALE \cite{akata2016label}, LatEm \cite{xian2016latent}, ESZSL \cite{romera2015embarrassingly}, SJE \cite{akata2015evaluation}, LESAE \cite{liu2018zero}, MFMR \cite{xu2017matrix}; (3) hybrid methods: UVDS \cite{long2017zero}, SPAEN \cite{chen2018zero}, DEM \cite{zhang2017learning}, SEZSL \cite{kumar2018generalized}.

\begin{table*}[ht]
	\centering
	\renewcommand\arraystretch{1.2}
	\caption{Generalized Zero-Shot Classification Accuracy (in \%) Using Proposed Split (PS). ts and tr are the Accuracy of Images from Unseen Classes and Seen Classes, Respectively. H is the Harmonic Mean Accuracy. Numbers in Brackets are Relative Performance Gains. Boldface Indicates the Best. `-' Indicates that no Reported Results are Available.}
	\label{table_gzsl}
	\setlength{\tabcolsep}{1.5mm}{
		\begin{tabular}{|c||ccc|ccc|ccc|ccc|ccc|}
			\hline
			\multirow{2}{*}{\textbf{Method}} &\multicolumn{3}{c|}{\textbf{AwA}} &\multicolumn{3}{c|}{\textbf{AwA2}} & \multicolumn{3}{c|}{\textbf{aPY}} & \multicolumn{3}{c|}{\textbf{CUB}} & \multicolumn{3}{c|}{\textbf{SUN}} \\
			& \textbf{ts}          & \textbf{tr}         & \textbf{H}   & \textbf{ts}          & \textbf{tr}         & \textbf{H}      & \textbf{ts}          & \textbf{tr}         & \textbf{H}& \textbf{ts}          & \textbf{tr}         & \textbf{H}& \textbf{ts}          & \textbf{tr}         & \textbf{H}       \\ \hline\hline
			ESZSL \cite{romera2015embarrassingly} &6.6 &75.6 &12.1 &5.9 &77.8 &11.0 &2.4 &70.1 &4.6 &12.6 &63.8 &21.0 &11.0 &27.9 &15.8      \\
			SJE \cite{akata2015evaluation} &11.3 &74.6 &19.6  &8.0 &73.9 &14.4 &3.7 &55.7 &6.9 &23.5 &59.2 &33.6 &14.7 &30.5 &19.8     \\
			SYNC \cite{changpinyo2016synthesized} &8.9 &87.3 &16.2 &10.0 &90.5 &18.0  &7.4 &66.3 &13.3 &11.5 &70.9 &19.8 &7.9 &43.3 &13.4    \\
			ALE \cite{akata2016label} & 16.8 &76.1 &27.5  &14.0 &81.8 &23.9 &4.6 &73.7 &8.7 &23.7 &62.8 &34.4 &21.8 &33.1 &26.3   \\
			LESAE \cite{liu2018zero} &19.1 &70.2 &30.0 &21.8 &70.6 &33.3 &12.7 &56.1 &20.1 &24.3 &53.0 &33.3 &21.9 &34.7 &26.9   \\
			SPAEN \cite{chen2018zero} &- &- &- &23.3 &90.9 &37.1  &13.7 &63.4 &22.6 &34.7 &70.6  &46.6 &24.9 &38.6 &30.3 \\ \hline
			DAP \cite{lampert2014attribute} &0.0 &88.7 &0.0 &0.0 &84.7 &0.0  &4.8 &78.3 &9.0 &1.7 &67.9 &3.3 &4.2 &25.1 &7.2    \\
			DAP+CA &20.6           &75.2          &32.3(+\textit{32.3})    &12.8 &71.2 &21.7(+\textit{21.7})        &6.0          &56.9           &10.8(+\textit{1.8})         &11.8         &38.3 &18.0(+\textit{14.7}) &5.0 &13.3 &7.3(+\textit{0.1})   \\ \hline
			LatEm \cite{xian2016latent} &7.3 &71.7 &13.3  &11.5 &77.3 &20.0 &0.1 &73.0 &0.2 &15.2 &57.3 &24.0 &14.7 &28.8 &19.5       \\
			LatEm+CA & 4.3 &86.4 &8.2(-\textit{5.1})  &13.2 &73.1 &22.4(+\textit{2.4}) &6.0 &83.1 &11.2(+\textit{11.0}) &14.3 &56.6 &22.8(-\textit{1.2}) &17.6 &28.2 &27.9(+\textit{8.4})  \\ \hline
			DEM \cite{zhang2017learning} &32.8 &84.7 &47.3 &30.5 &86.4 &45.1 &11.1 &75.1 &19.4 &19.6 &57.9 &29.2 &20.5 &34.3 &25.6   \\
			DEM+CA &35.6  &86.3 &50.4(+\textit{3.1}) &32.2 &85.9 &46.9(+\textit{1.8})  &14.2 &78.5 &\textbf{24.0}(+\textit{4.6}) &20.4 &60.3 &30.5(+\textit{1.3}) &22.8  &39.2 &28.8(+\textit{3.2})  \\ \hline
			MFMR \cite{xu2017matrix} &11.7 &78.5 &20.3 &17.7 &77.2 &28.8  &8.5 &71.4 &15.1 &8.4 &13.9 &10.5 &3.7 &8.1 &5.1  \\
			MFMR+CA & 12.3 &78.6 &21.2(+\textit{0.9}) &18.4 &82.4 &30.1(+\textit{1.3})  &9.7 &73.9 &17.2(+\textit{2.1}) &9.2 &14.3 &11.2(+\textit{0.7}) &4.0 &7.2 &5.2(+\textit{0.1})  \\ \hline
			SEZSL \cite{kumar2018generalized}  &56.3 &67.8 &61.5 &58.3 &68.1 &62.8 &- &- &- &41.5 &53.3 &46.7 &40.9 &30.5 &34.9    \\
			SEZSL+CA &58.2 &70.4 &\textbf{63.7}(+\textit{2.2}) &60.1 &67.9 &\textbf{63.8}(+\textit{1.0}) &- &- &- &44.7 &56.1 &\textbf{49.8}(+\textit{3.1}) &39.6 &35.5 &\textbf{37.4}(+\textit{2.5})   \\ \hline		
	\end{tabular}}
\end{table*}

\subsection{Comparison with the State-of-the-Art}

\subsubsection{Conventional Zero-shot Learning}

To evaluate the improvement that benefits from the complementary attributes, we modify five attribute-based ZSL methods (i.e. DAP, LatEm, DEM, MFMR and SEZSL) by adopting CA, and test them on five benchmark datasets. The comparison results are summarized in Table \ref{table_zsl}, where the numbers in brackets are the relative improvements after using CA. It can be observed that CA can significantly improve the performance of existing ZSL methods (except for LatEm on AwA dataset\footnote{This is because the result of the original LatEm is cited from \cite{xian2018zero}, which is higher than what we have reproduced. In fact, the performance of the improved LatEm by CA is better than that of our reimplemented LatEm.}). Specifically, the mean class accuracies of five ZSL methods are increased by 6.83\%, 3.01\%, 1.88\%, 1.19\% and 1.83\% (2.95\% on average) after using complementary attributes. Moreover, there are 9 out 43 method-dataset combinations achieve more than 5\% improvements, and 25 out of 43 achieve more than 2\% improvements. These results suggest that the proposed complementary attributes are consistently effective to improve existing ZSL methods. In addition, CA increases the classification accuracy by 6.83\% for the probability-prediction strategy based ZSL method (e.g. DAP), and by 3.01\% for the label-embedding strategy (e.g. LatEm). These results are explicable since that the improvement of the probability-prediction strategy benefits from both the complementary attributes and the rank aggregation. 

Moreover, we compare the modified ZSL methods by complementary attributes (ZSLCA) with several state-of-the-art baselines on five datasets with both standard split and proposed split. In Table \ref{table_zsl}, we can observe that the modified ZSL methods outperform the original methods, and also have a significant advantage comparing to the state-of-the-art on all datasets. The best results on five datasets (AwA, AwA2, aPY, CUB and SUN) using PS are increased by 1.2\%, 1.7\%, 0.9\%, 2.7\% and 2.1\%, respectively (1.7\% on average) after adopting CA, which demonstrates the effectiveness and robustness of the complementary attributes.

\subsubsection{Generalized Zero-shot Learning}
Zero-shot learning has a restrictive set up as it comes with a strong assumption that the images tested at test time can only come from unseen classes. As a more realistic setting, generalized zero-shot learning \cite{xian2018zero} has been proposed to generalize the conventional ZSL task to the case that test images can be derived from both unseen and seen classes. We evaluate complementary attributes with the generalized ZSL setting on five datasets and calculate the harmonic mean accuracy as the evaluation criteria. The experimental results including the classification accuracies of unseen classes and seen classes and harmonic mean accuracy are shown in Table \ref{table_gzsl}. It can be observed that the harmonic mean accuracy of DAP has a significant increase (14.12\% on average) after using complementary attributes. Although the performance of original DAP is very terrible on AwA dataset, CA and RA can improve it to an acceptable level. These results demonstrate that the proposed complementary attributes and rank aggregation can enhance the representation ability of semantic attributes, and consequently improve the performance of existing ZSL methods on the generalized ZSL task. 

\begin{table}[t]
	\centering
	\renewcommand\arraystretch{1.2}
	\caption{Classification Accuracy (in \%) on ImageNet Dataset. Numbers in Brackets are Relative Performance Gains. Boldface Indicates the Best.}
	\label{table_imagenet}
		\setlength{\tabcolsep}{2mm}{	
			\begin{tabular}{|c||ccc|}
				\hline
				\textbf{Method} &\textbf{2-hop} &\textbf{3-hop} &\textbf{All 20K} \\ \hline \hline
				LatEm \cite{xian2016latent} &5.45 &1.32 &0.50 \\
				LatEm+CA &5.93(+\textit{0.48}) &1.51(+\textit{0.19}) &0.57(+\textit{0.07}) \\\hline
				ESZSL \cite{romera2015embarrassingly} &6.35 &1.51 &0.62 \\
				ESZSL+CA &6.84(+\textit{0.49}) &1.62(+\textit{0.11}) &0.70(+\textit{0.08}) \\\hline
				SJE \cite{akata2015evaluation} &5.31 &1.33 &0.52 \\
				SJE+CA &5.88(+\textit{0.57}) &1.45(+\textit{0.12}) &0.60(+\textit{0.08}) \\ \hline 			
				SYNC \cite{changpinyo2016synthesized} &9.26 &2.29 &0.96 \\
				SYNC+CA &\textbf{9.53}(+\textit{0.27}) &\textbf{2.32}(+\textit{0.03}) &\textbf{0.98}(+\textit{0.02}) \\\hline
	\end{tabular}}		
\end{table}

\subsubsection{Results on ImageNet}
Complementary attributes are evaluated on the large-scale ImageNet dataset and the results are shown in Table \ref{table_imagenet}. It is obvious that after adopting the complementary attributes, existing ZSL methods could make great progress. Specifically, the average accuracies of the four ZSL methods are increased by $ 0.45\% $ and $ 0.11\% $ on 2-hop and 3-hop setting respectively. Even on the evaluation of all 20K class, the average classification accuracy is improved by $ 0.06\% $. Experimental results demonstrate the efficacy of complementary attributes for zero-shot classification on the large-scale dataset.

\subsection{Ablation Study}

\subsubsection{Evaluation of Complementary Attributes}
\begin{table}[t]
	\centering
	\renewcommand\arraystretch{1.2}
	\caption{Classification Accuracy (in \%) Including Mean Class Accuracy (MC acc.) and Mean Sample Accuracy (MS acc.) on AwA Dataset. Boldface Indicates the Best.}
	\label{table_dap}
	\setlength{\tabcolsep}{1.5mm}{	
		\begin{tabular}{|c||cc|cc|}
			\hline
			\multirow{2}{*}{\textbf{Method}} & \multicolumn{2}{c|}{\textbf{SS}} & \multicolumn{2}{c|}{\textbf{PS}} \\ 
			&\textbf{MC acc.}  &\textbf{MS acc.}  &\textbf{MC acc.}  &\textbf{MS acc.} \\ \hline \hline
			{DAP} &57.1 &60.9 &44.1 &40.8 \\ 
			{DAP+CA} &68.8 &65.7 &52.4 &51.2 \\ 
			{DAP+RA} &66.4 &63.7 &50.5 &46.9 \\ 
			{DAP+CA+RA} &\textbf{69.7} &\textbf{66.7} &\textbf{54.4} &\textbf{54.1} \\ 
			\hline
	\end{tabular}}		
\end{table}

In the first experiment, we evaluate the improvement that benefits from the complementary attributes. We compare the proposed method with DAP on AwA dataset using both SS and PS. The classification accuracies (including mean class accuracy and mean sample accuracy) of the original DAP and the improved DAP with complementary attributes (DAPCA) are presented in Table \ref{table_dap}. It can be observed that adopting complementary attributes can significantly improve the performance of the original ZSL model. Complementary attributes, as the supplement to the original attributes, can describe objects more comprehensively and consequently make the semantic embedding space more complete. The experimental results show that DAPCA can improve the original DAP by 11.7\% and 8.3\% on SS and PS respectively. These experimental results demonstrate that complementary attributes are an effective supplement to the original attributes that can significantly improve the performance of existing ZSL methods.

\begin{figure}[t]
	\centering
	\subfigure[Original DAP]{
		\label{figure_confusionmatrix_1}
		\includegraphics[width = 0.22\textwidth]{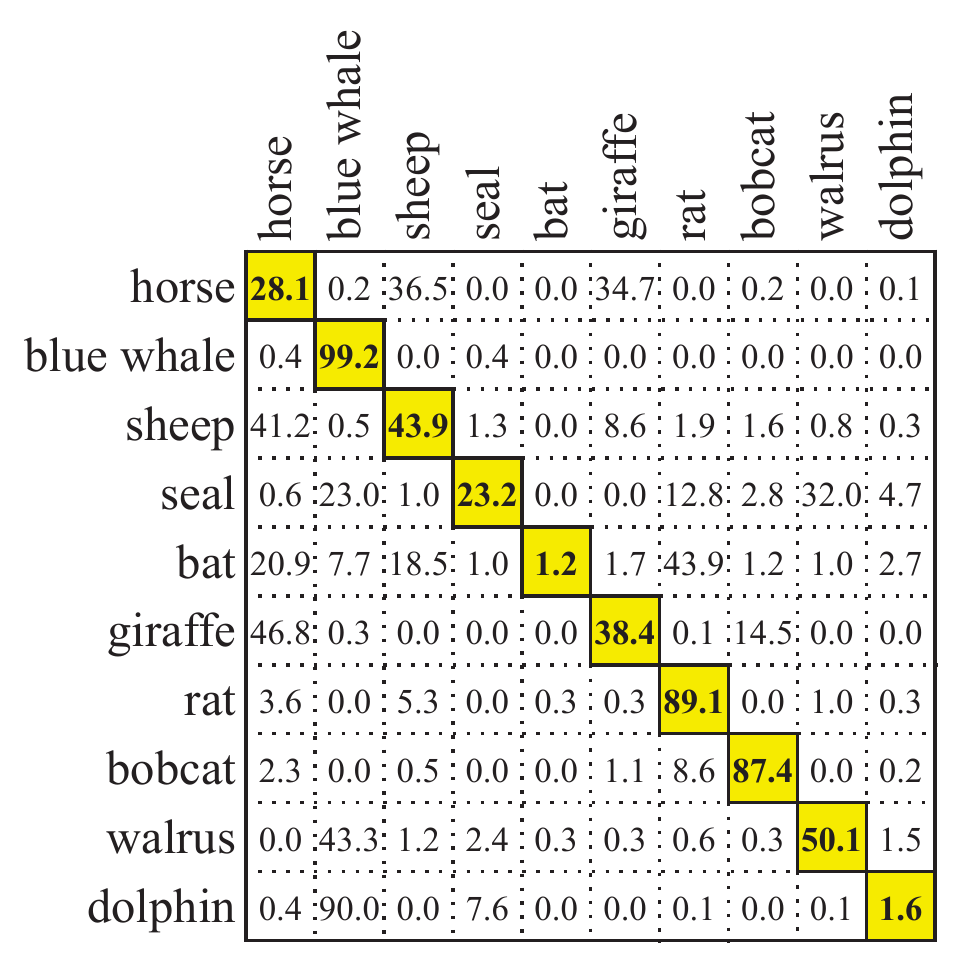}
	}
	\subfigure[DAP+CA]{
		\label{figure_confusionmatrix_2}
		\includegraphics[width = 0.22\textwidth]{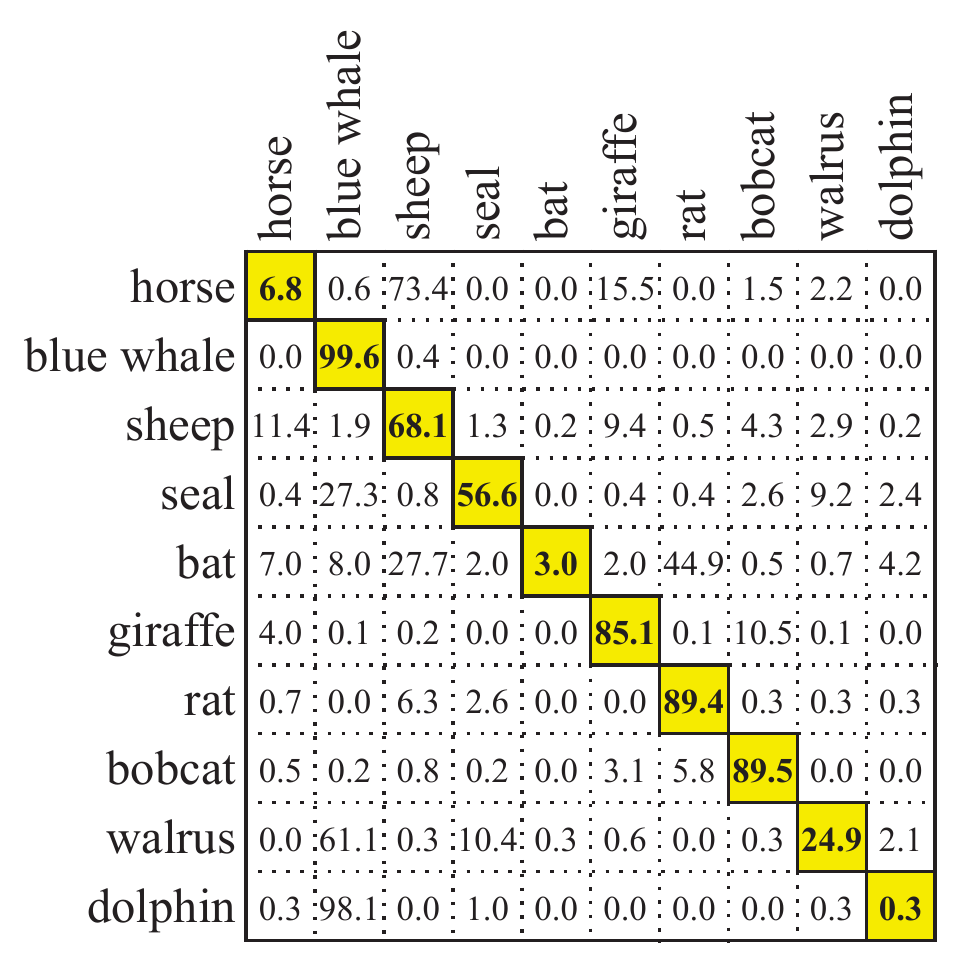}
	}
	\subfigure[DAP+RA]{
		\label{figure_confusionmatrix_3}
		\includegraphics[width = 0.22\textwidth]{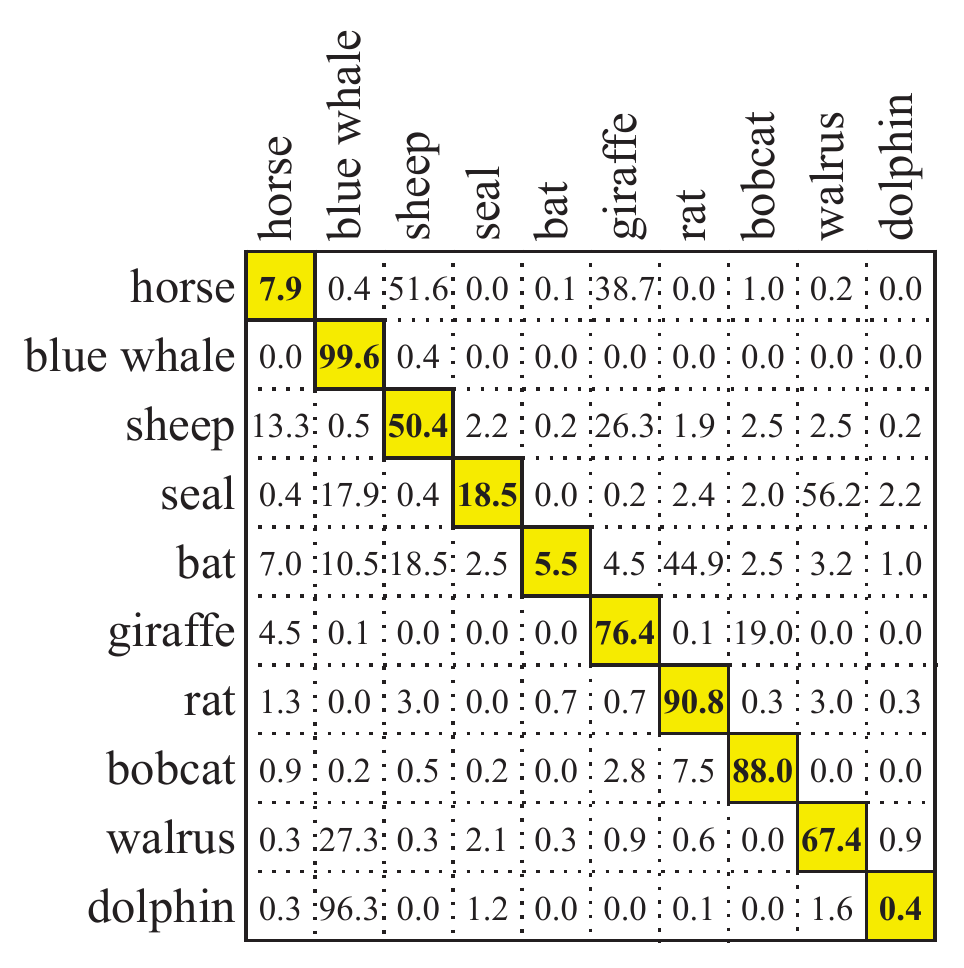}
	}
	\subfigure[DAP+CA+RA]{
		\label{figure_confusionmatrix_4}
		\includegraphics[width = 0.22\textwidth]{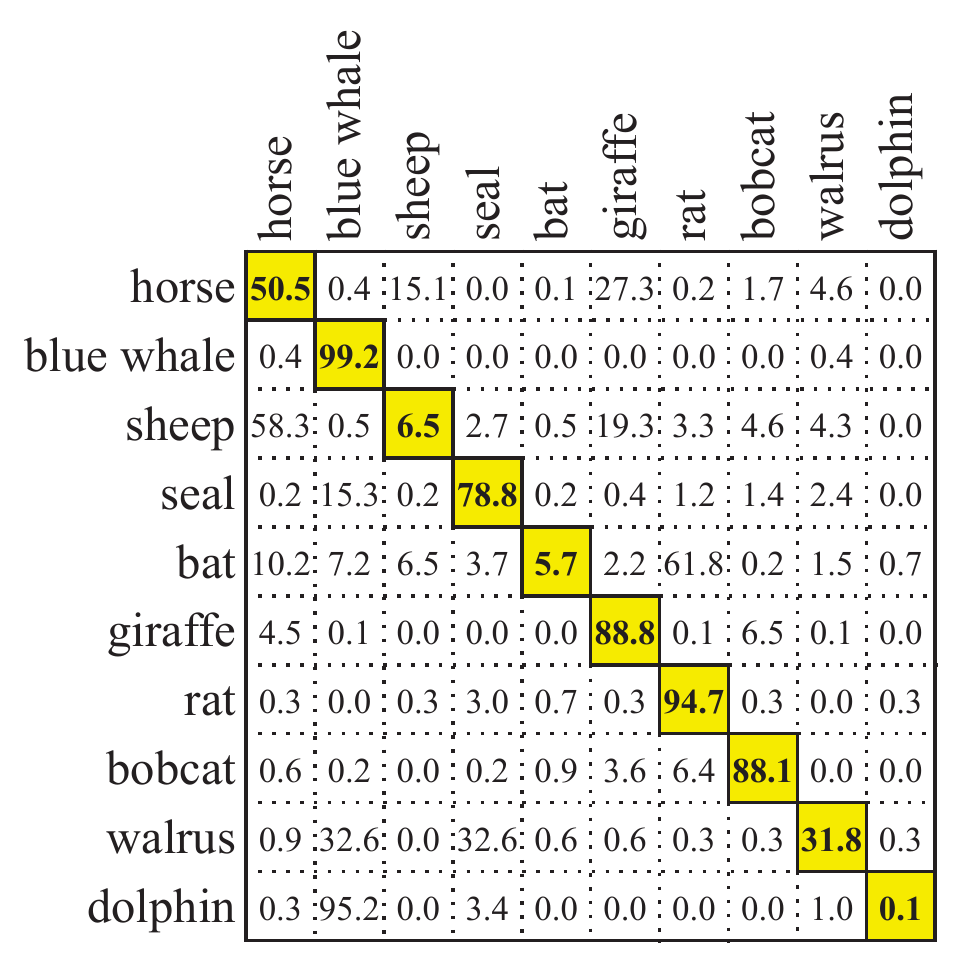}
	}
	\caption{Confusion matrices (in \%) between 10 test classes on AwA dataset using PS. (a) Original DAP; (b) DAP with complementary attributes; (c) DAP with rank aggregation; (d) DAP with both complementary attributes and rank aggregation.}
	\label{figure_confusionmatrix}
\end{figure}

In order to analyze the robustness of adopting complementary attributes, confusion matrices of the original DAP and DAPCA on AwA dataset using proposed split are illustrated in Fig. \ref{figure_confusionmatrix}. The numbers in the diagonal area (yellow patches) of confusion matrices indicate the classification accuracy per class. From Fig. \ref{figure_confusionmatrix_1} and Fig. \ref{figure_confusionmatrix_2}, we can see that DAPCA performs better on most of the test classes, and the accuracy of DAPCA nearly doubles the original DAP's on some test classes, such as \textit{seal} and \textit{giraffe}. And more importantly, the confusion matrix of DAPCA contains less noise (i.e. smaller numbers in the side regions (white patches) of confusion matrices) than DAP’s, which suggests that DAPCA has less prediction uncertainties. In other words, complementary attributes can significantly and robustly improve the performance of existing attribute-based ZSL methods.

\begin{figure}[t]
	\centering 
	\subfigure[Original Attributes]{ 
		\label{fig_tsne_a} 
		\includegraphics[width=0.22\textwidth]{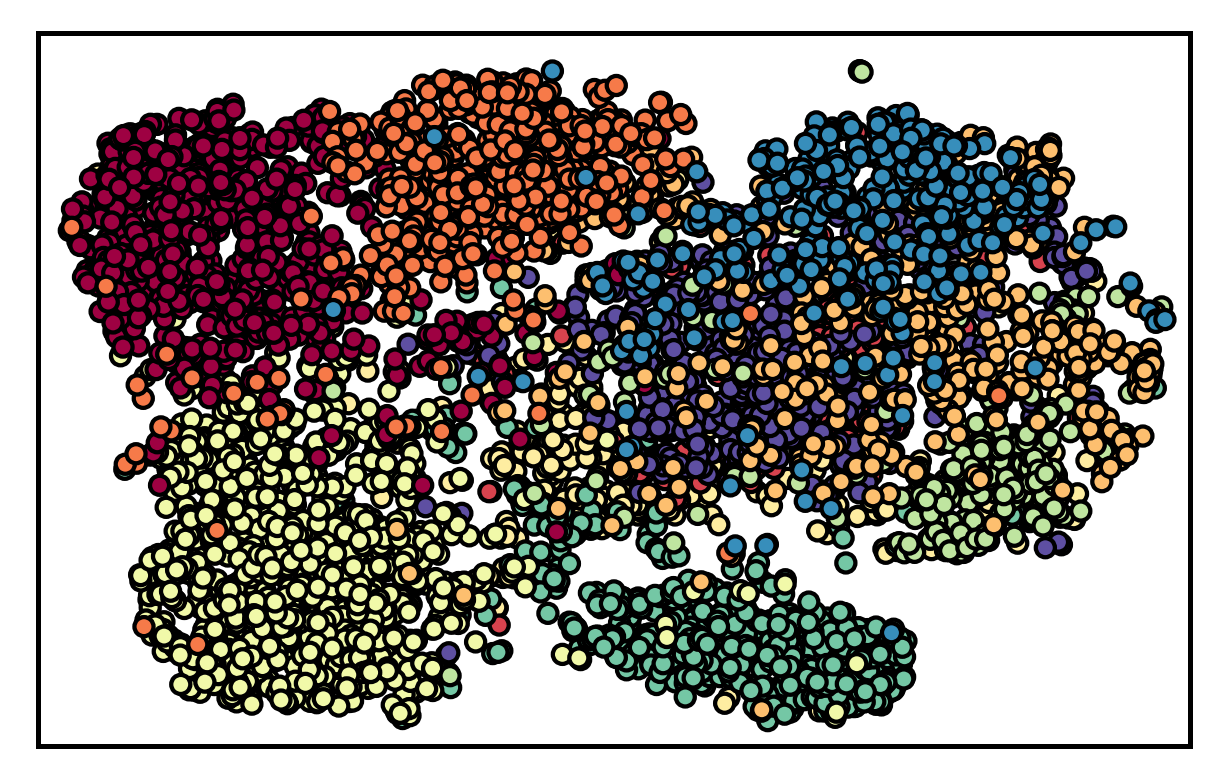} 
	} 
	\subfigure[Complementary Attributes]{ 
		\label{fig_tsne_b}
		\includegraphics[width=0.22\textwidth]{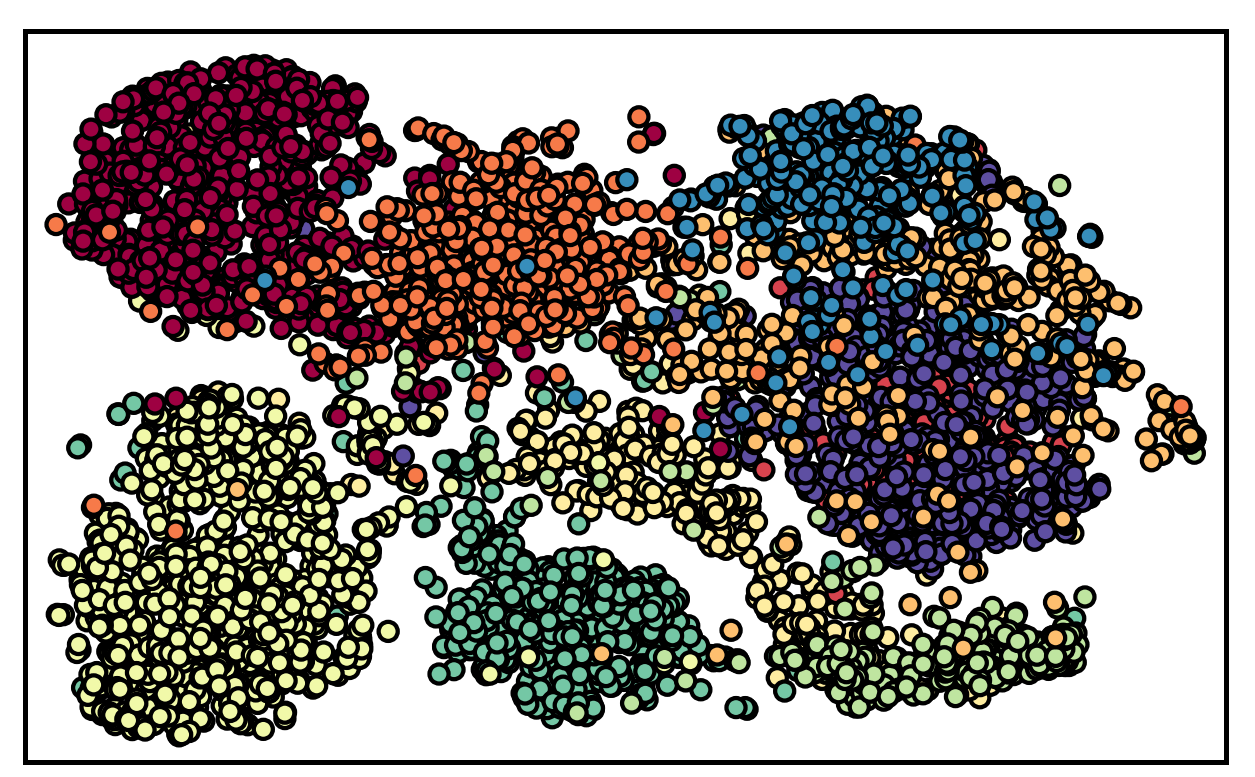} 
	} 
	\caption{T-SNE of semantic representation spaces of LatEm on AwA dataset. (a) original attributes, (b) complementary attributes.} 
	\label{fig_tsne}
\end{figure}

In the second experiment, we visualize the distribution of the semantic representation spaces of original attributes and complementary attributes using T-SNE \cite{maaten2008visualizing} in Fig. \ref{fig_tsne}. The semantic spaces are generated on AwA dataset using original LatEm and LatEm with complementary attributes (LatEmCA), respectively. It is obvious that, test samples projected in the complementary attributes space (Fig. \ref{fig_tsne}(b)) are more discriminative comparing to the distribution in the original attributes space (Fig. \ref{fig_tsne}(a)). The T-SNE result is another evidence suggests that complementary attributes are a supplement to the original attributes which can make a more discriminative semantic space.

\begin{figure*}[t]
	\centering 
	\subfigure[DAP, AwA]{ 
		\label{fig_subset_d_1}
		\includegraphics[width=0.18\textwidth]{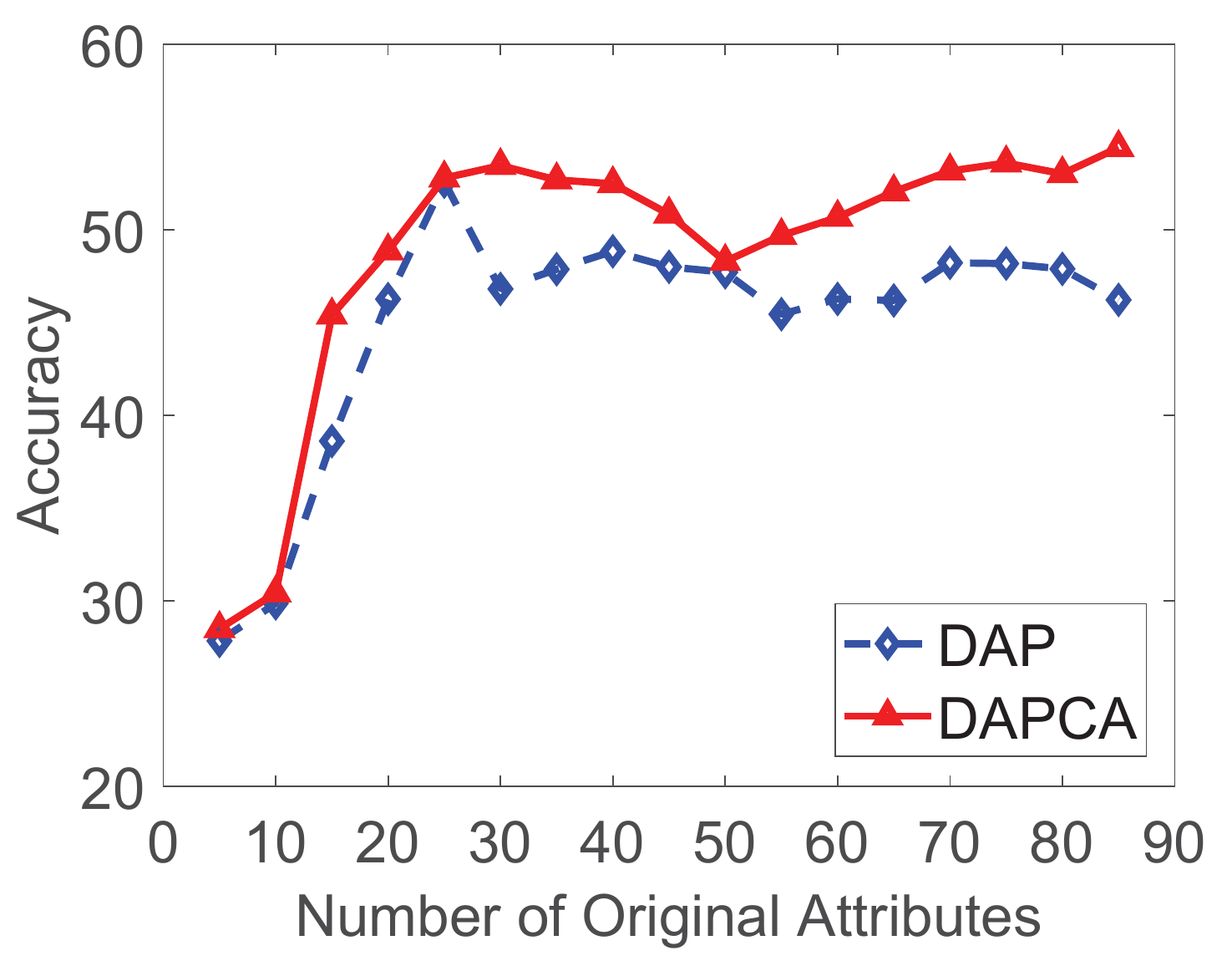} 
	} 
	\subfigure[DAP, AwA2]{ 
		\label{fig_subset_d_2} 
		\includegraphics[width=0.18\textwidth]{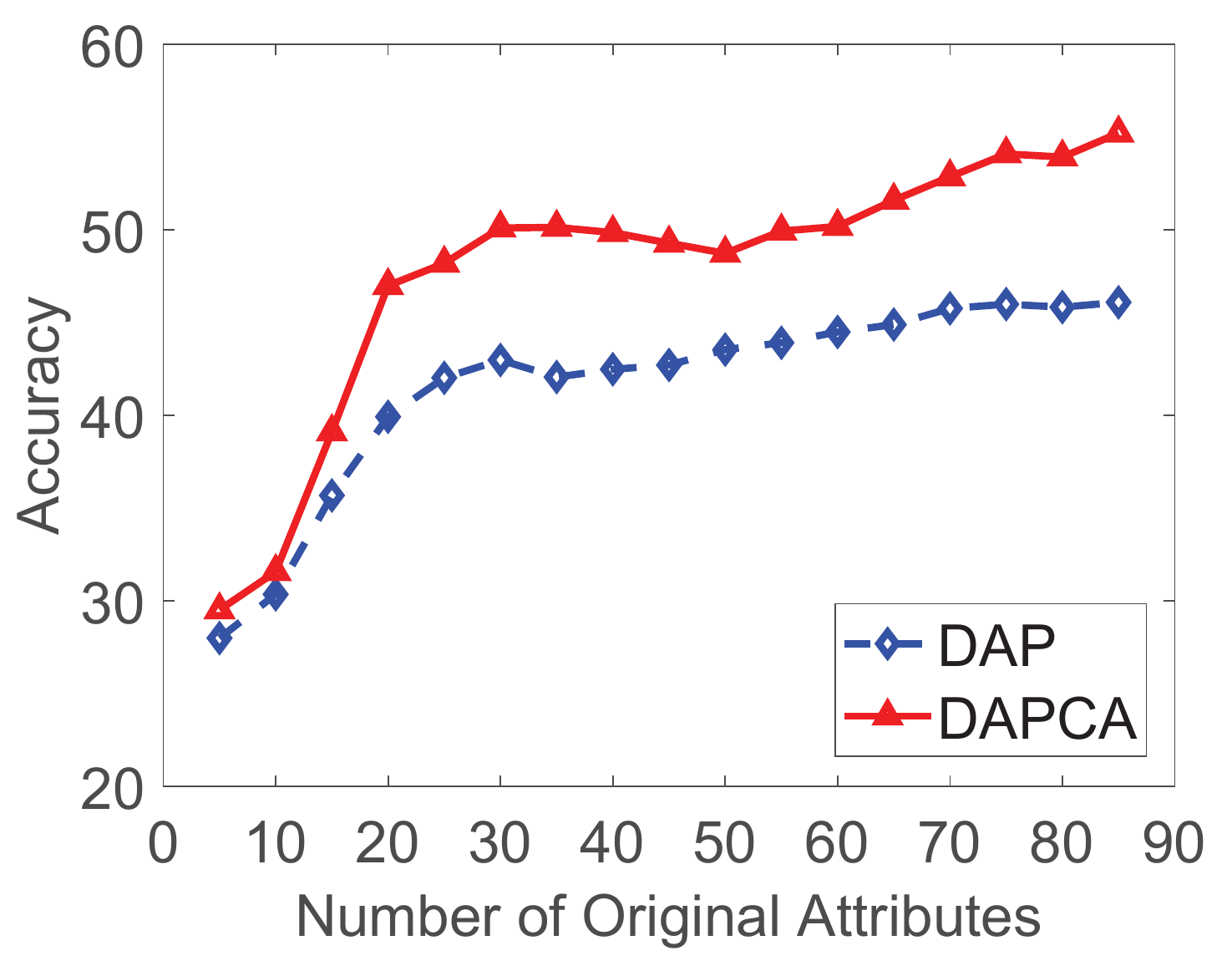} 
	} 
	\subfigure[DAP, aPY]{ 
		\label{fig_subset_d_3}
		\includegraphics[width=0.18\textwidth]{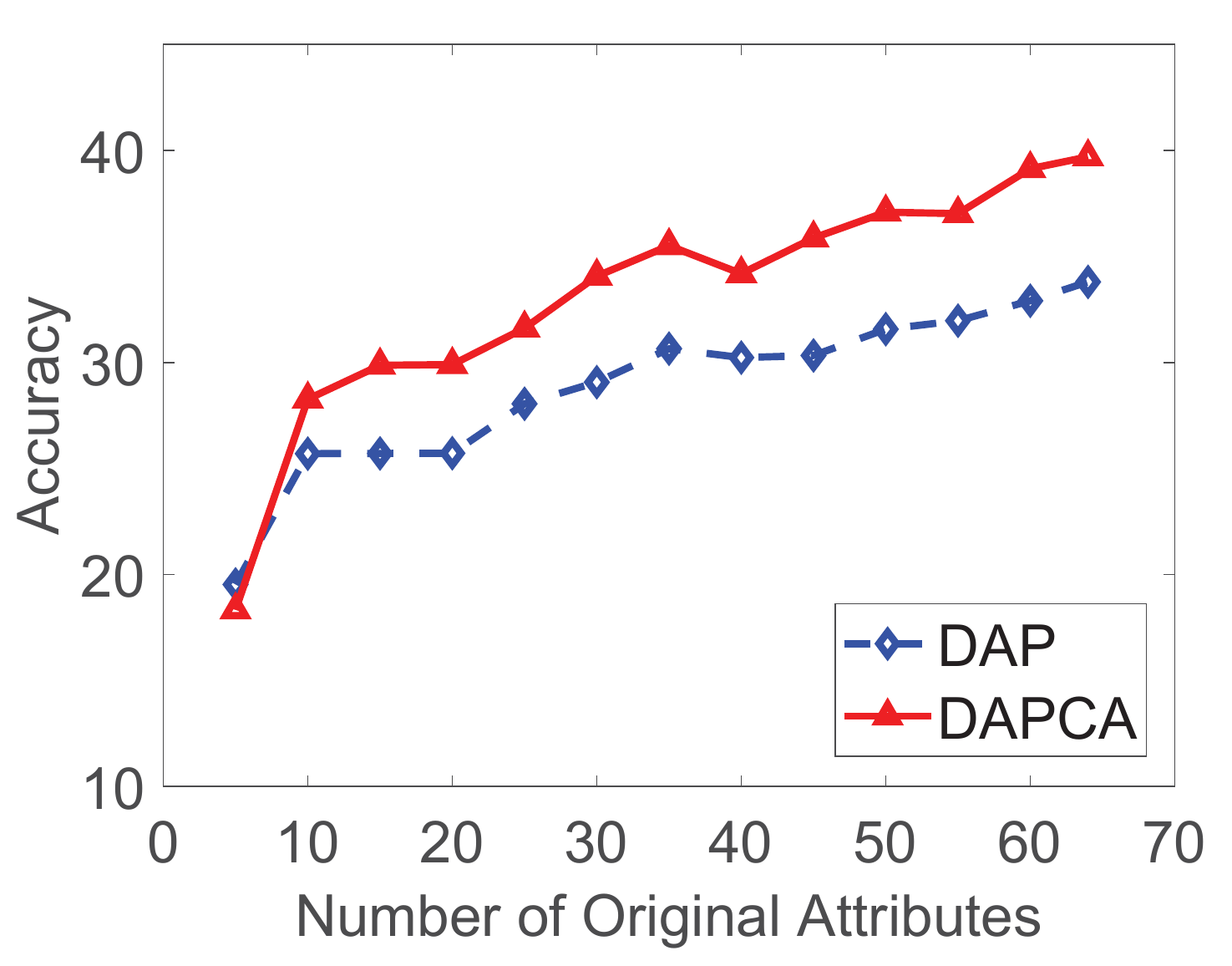} 
	} 
	\subfigure[DAP, CUB]{ 
		\label{fig_subset_d_4}
		\includegraphics[width=0.18\textwidth]{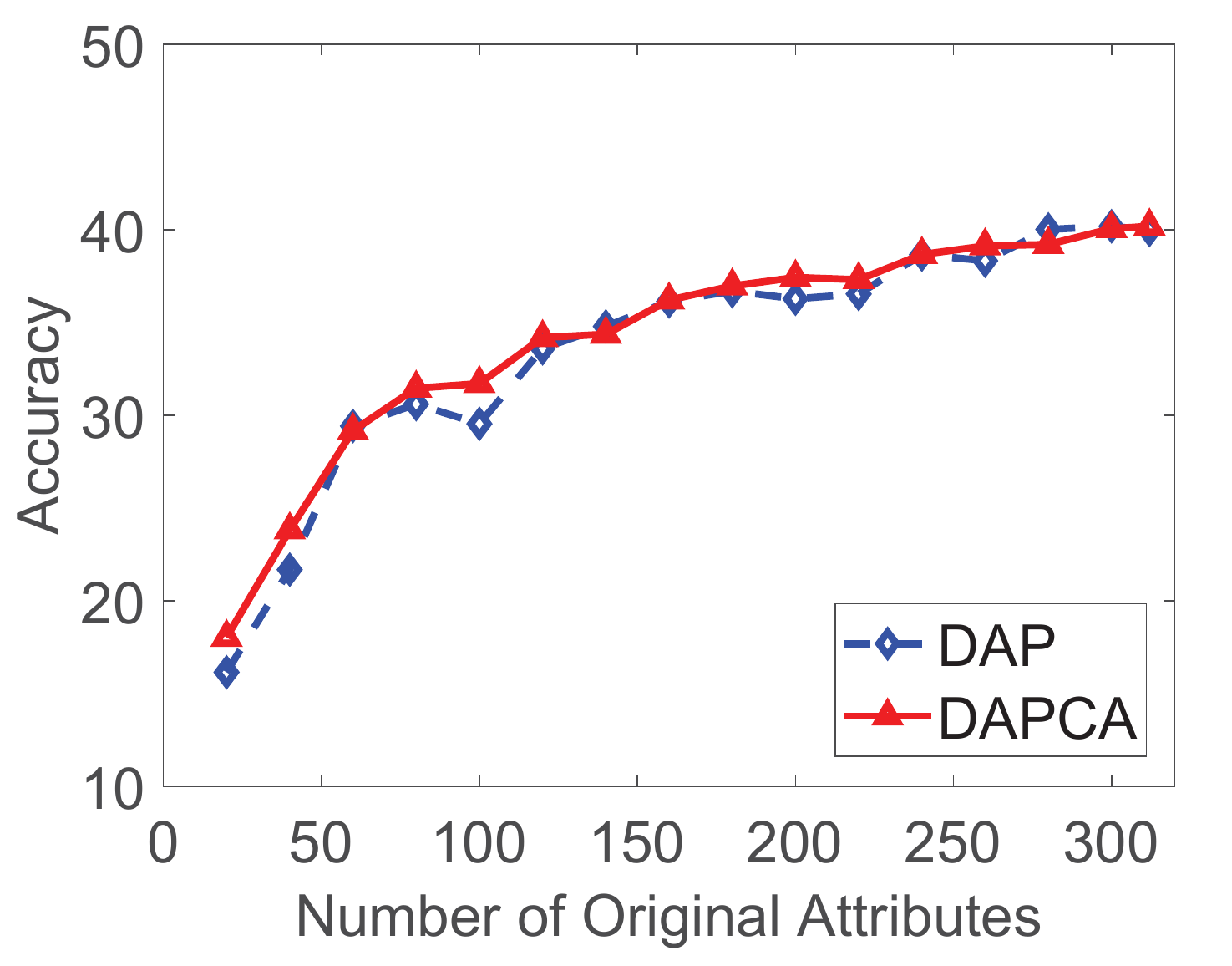} 
	} 
	\subfigure[DAP, SUN]{ 
		\label{fig_subset_d_5}
		\includegraphics[width=0.18\textwidth]{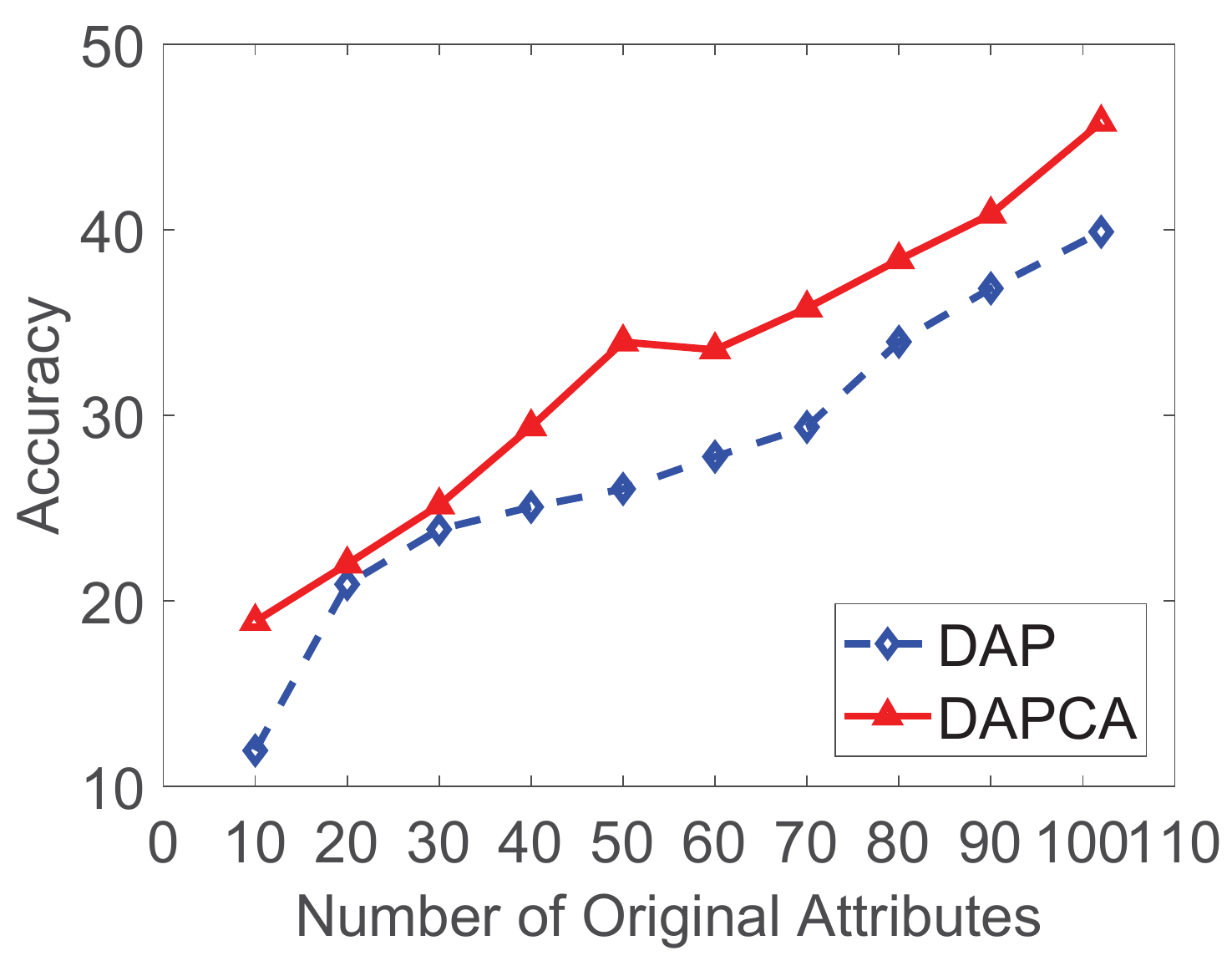} 
	} 
	\subfigure[LatEm, AwA]{ 
		\label{fig_subset_l_1}
		\includegraphics[width=0.18\textwidth]{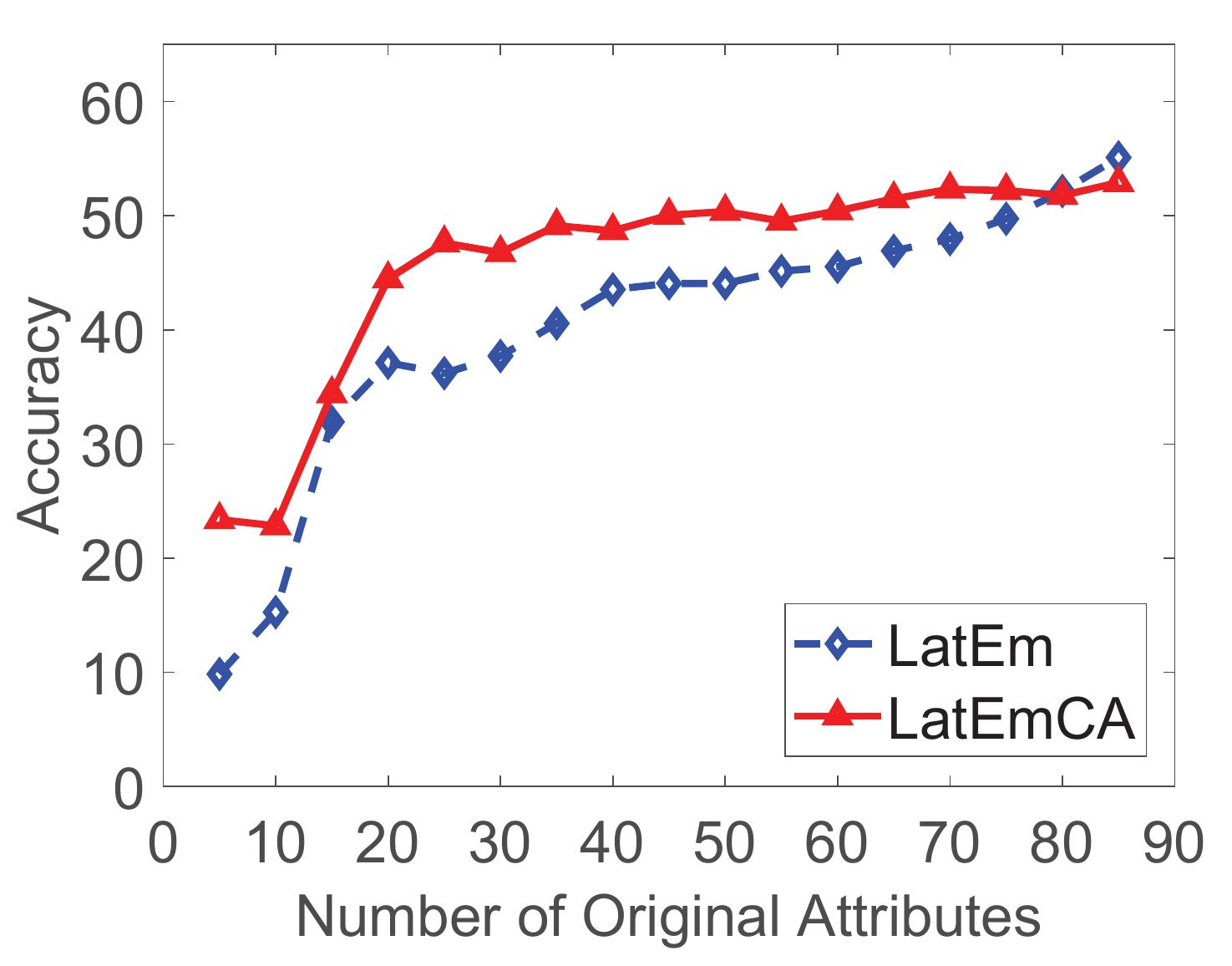} 
	} 
	\subfigure[LatEm, AwA2]{ 
		\label{fig_subset_l_2}
		\includegraphics[width=0.18\textwidth]{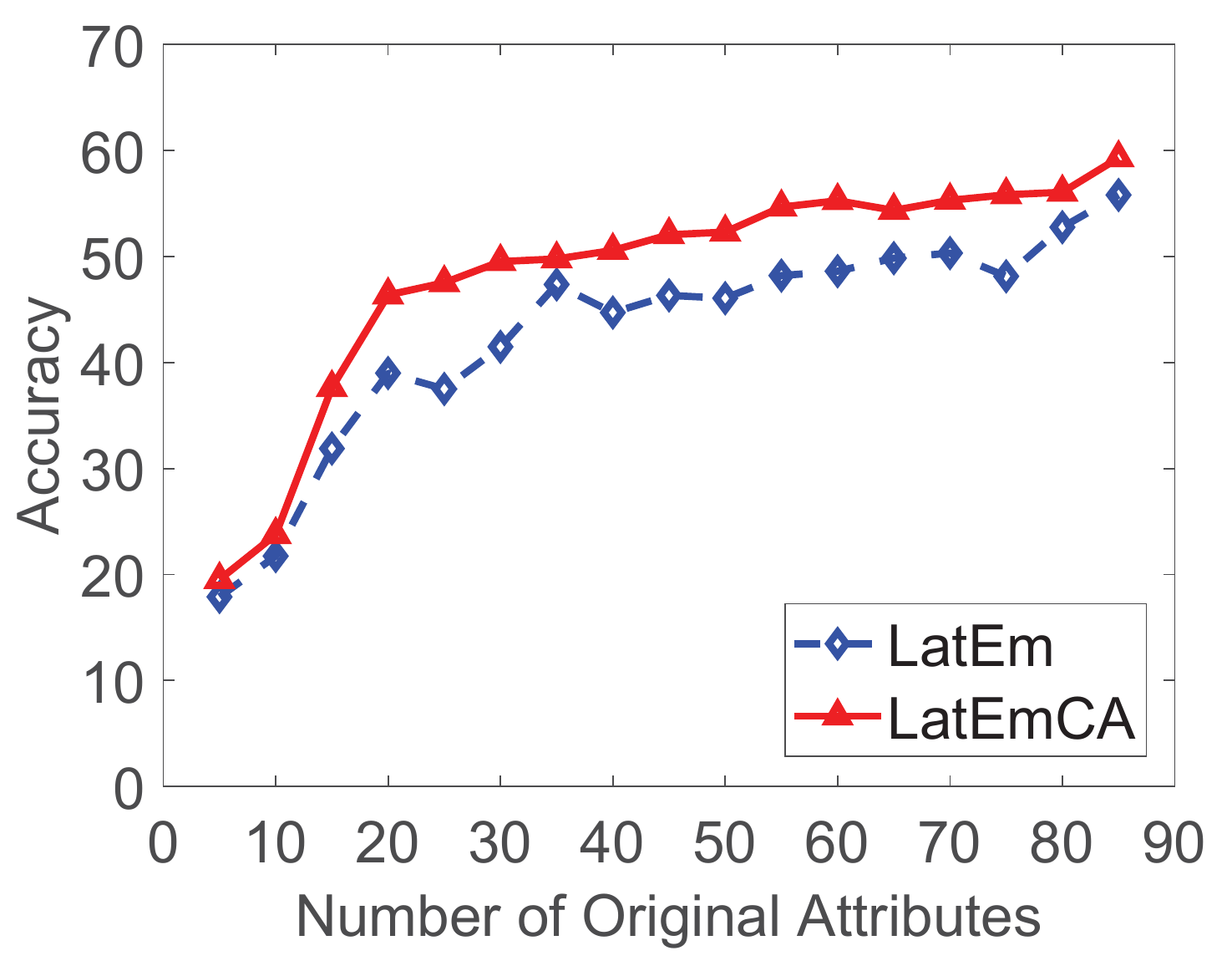} 
	} 
	\subfigure[LatEm, aPY]{ 
		\label{fig_subset_l_3} 
		\includegraphics[width=0.18\textwidth]{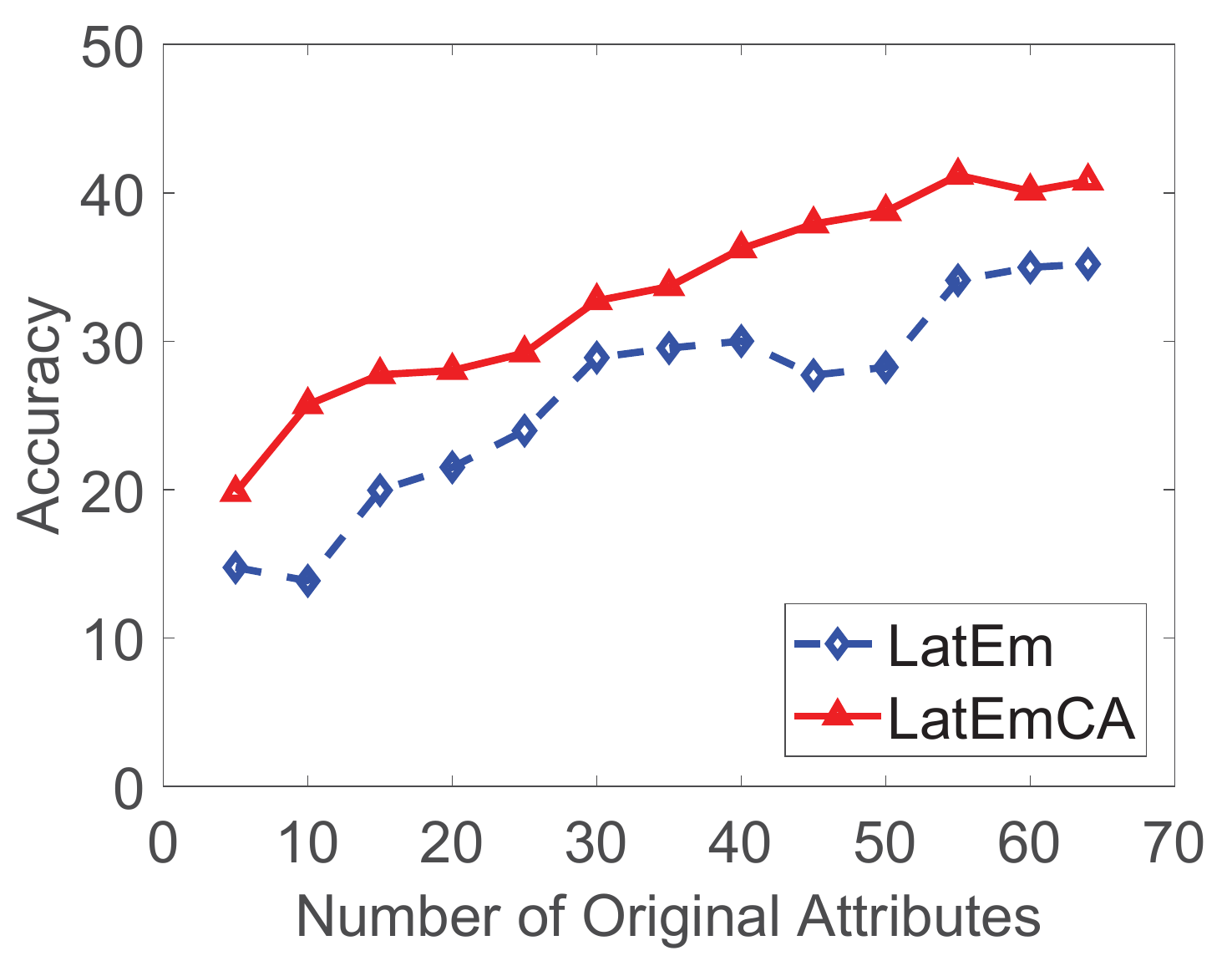} 
	} 
	\subfigure[LatEm, CUB]{ 
		\label{fig_subset_l_4} 
		\includegraphics[width=0.18\textwidth]{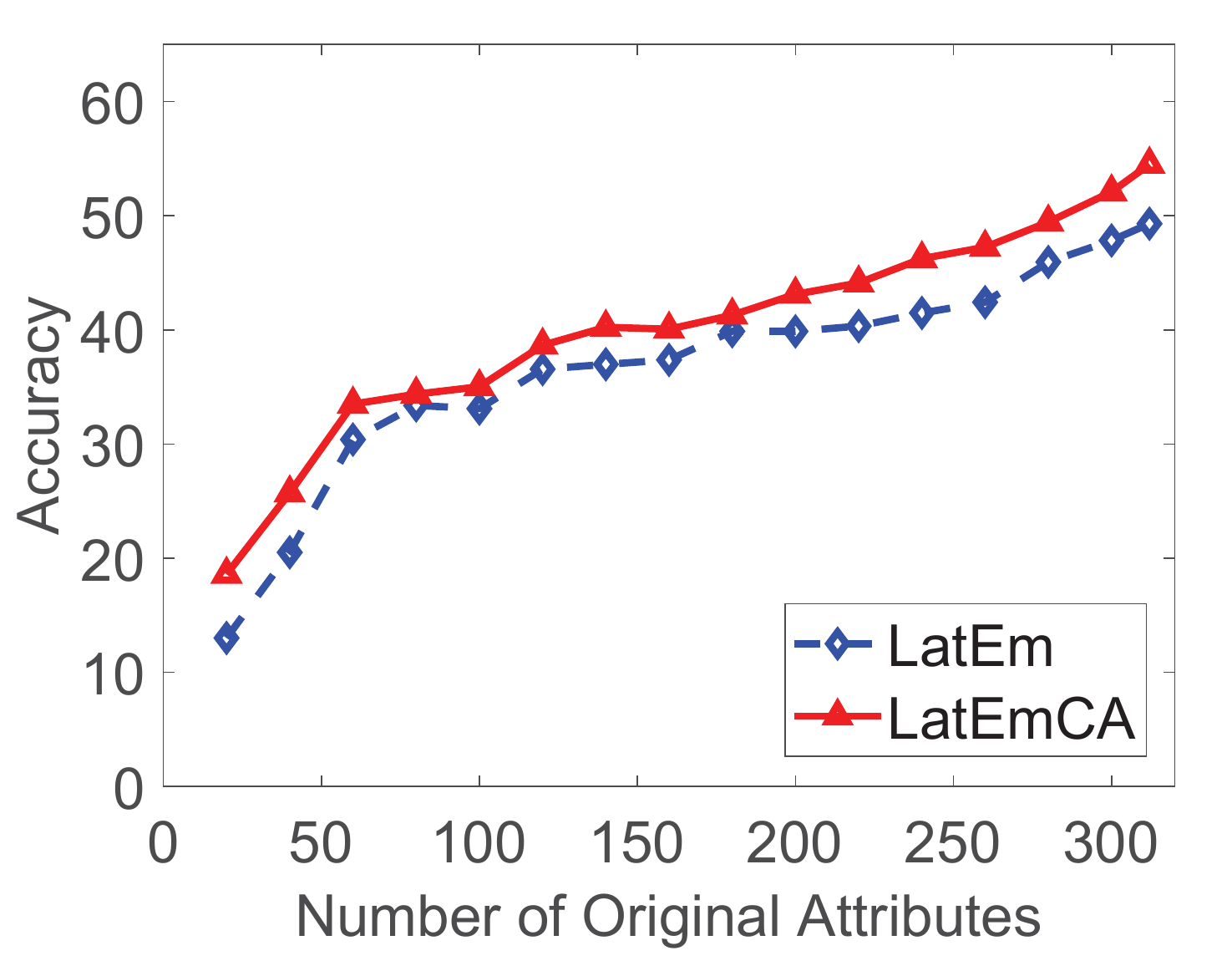} 
	} 
	\subfigure[LatEm, SUN]{ 
		\label{fig_subset_l_5} 
		\includegraphics[width=0.18\textwidth]{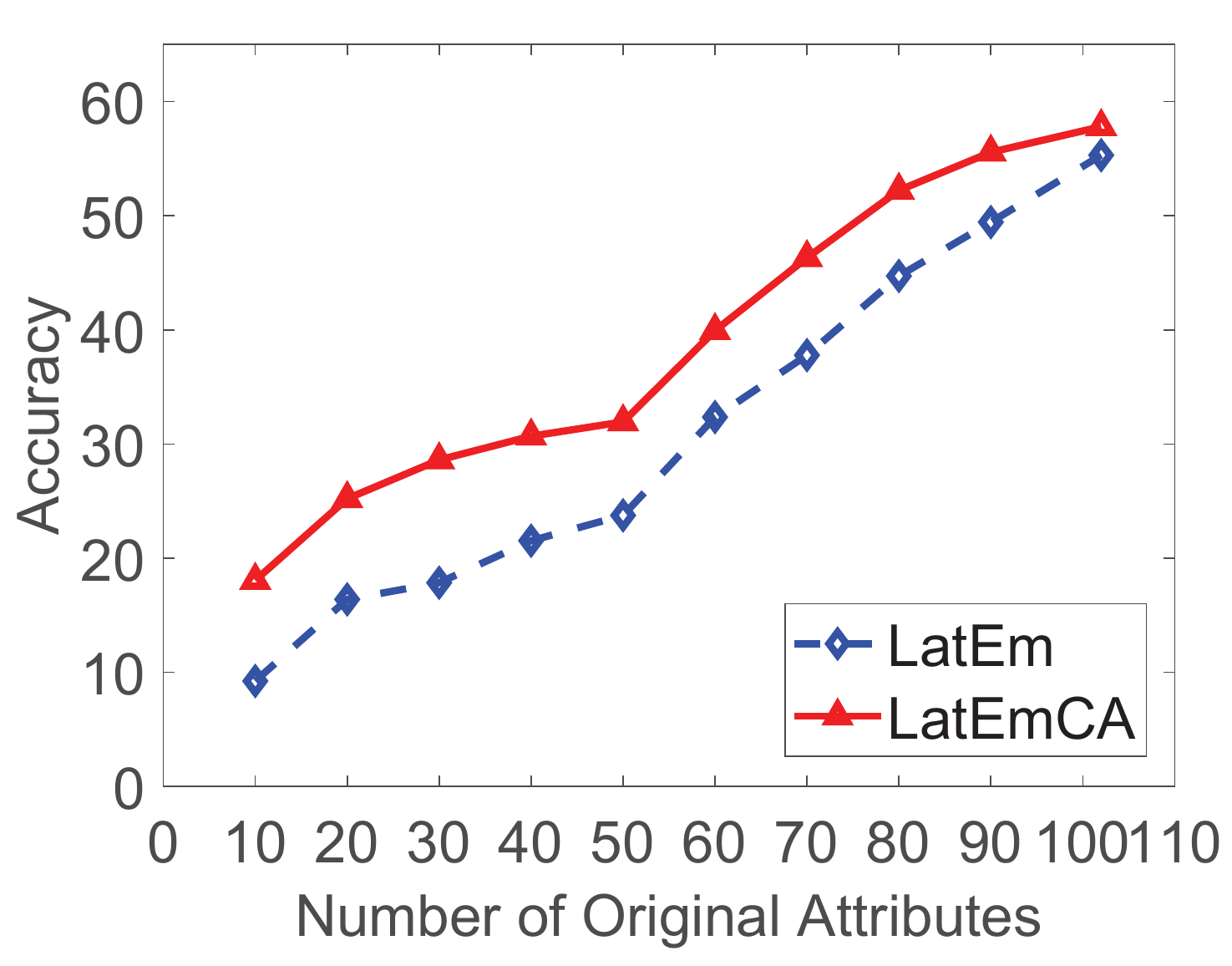} 
	} 
	
	\caption{The accuracy (in \%) with respect to the number of original attributes for DAP and LatEm on five datasets using PS. (a-j): Method, Dataset.} 
	\label{fig_subset}
\end{figure*}

In the third experiment, we compare ZSL with complementary attributes (ZSLCA) to the original ZSL methods on the setting that the number of used attributes changes. The accuracy with respect to the number of used attributes is shown in Fig. \ref{fig_subset}. We can see that, the performance of ZSLCA is greater than original ZSL method regardless of the number of used attributes. These results demonstrate that the improvement of ZSLCA is mainly due to the leverage of complementary attributes rather than the increase in the number of attributes. In other words, complementary attributes can make all the original attributes significant and help to represent a more comprehensive semantic space.

\subsubsection{Evaluation of Rank Aggregation}
To evaluate the efficacy of rank aggregation, we compare DAP with rank aggregation (DAPRA) to original DAP on AwA dataset using both SS and PS. The comparison results are shown in Table \ref{table_dap}. It can be observed that, comparing to DAP, DAPRA can improve the classification accuracy by 9.3\% and 6.4\% on standard split and proposed split respectively. These results show that the proposed rank aggregation framework is effective to improve the original DAP. This observation is comprehensible since the rank aggregation framework overcomes the slack assumption in PPZSL that attributes are independent of each other. Furthermore, although complementary attributes improve the performance of DAP, the rank aggregation framework can further get a performance gain based on DAPCA. 

We also show the confusion matrix of DAPRA and DAP with both complementary attributes and rank aggregation (DAPCR) in Fig. \ref{figure_confusionmatrix}. It is obvious that DAPRA performs better than DAP on most test classes, and the accuracy of DAPRA nearly doubles the original DAP's on class \textit{giraffe}. After using both complementary attributes and rank aggregation, the classification accuracies are improved to an acceptable level for some test classes whose performance is terrible on original DAP, such as class \textit{seal} (78.8\% vs 23.2\%) and \textit{giraffe} (88.8\% vs 38.4\%). These experimental results demonstrate that both complementary attributes and rank aggregation are effective and robust to improve existing ZSL methods.

\section{Conclusion}
\label{sec_conclu}
Zero-shot learning is a challenging but realistic task that aims to recognize new unseen objects using disjoint seen objects. To solve the problem of domain shifting, high-level semantic representation is adopted to transfer knowledge from seen domain to unseen domain. In this paper, we introduce complementary attributes, as the supplement to the original attributes, to help describe objects more comprehensively. Complementary attributes can be easily applied to various attribute-based ZSL methods, including the probability-prediction strategy based ZSL model and the label-embedding strategy based ZSL model. Moreover, we theoretically prove that adopting CA can improve the generalization bound of original ZSL methods. For the probability-prediction strategy based ZSL model, we propose the rank aggregation framework to circumvent the assumption that attributes are independent of each other. We conduct extensive experiments to evaluate the efficacy of complementary attributes and rank aggregation. Experimental results demonstrate that complementary attributes and rank aggregation can effectively and robustly improve the performance of existing ZSL methods.

This work is concentrated on the attributes. Obviously, other assistant information, which can be formalized to a relationship matrix as shown in \eqref{eq_A}, can be easily expanded by the corresponding complementary formulation. In the future, we plan to fuse multi-sources assistant information to improve the discriminative ability of current semantic representation.

\bibliographystyle{IEEEtran}
\bibliography{IEEEabrv,IEEEexample}
\vspace{-15 mm} 
\begin{IEEEbiography}[{\includegraphics[width=1in,height=1.25in,clip,keepaspectratio]{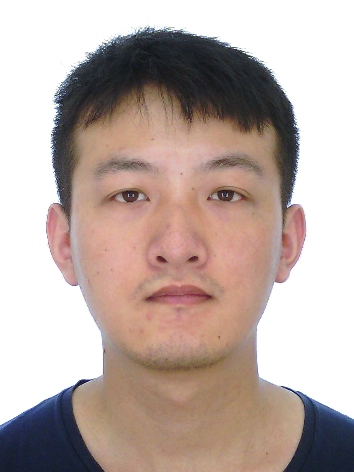}}]{Xiaofeng Xu}
	received the B.S. degree in computer science from Nanjing University of Science and Technology in 2014, where he is currently pursuing the PhD degree with the School of Computer Science and Engineering. From 2017 to now, he is the visiting research student with the Centre for Artificial Intelligence, University of Technology Sydney. His current research interests include machine learning, deep learning, zero-shot learning and computer vision.
\end{IEEEbiography}
\vspace{-15 mm} 
\begin{IEEEbiography}[{\includegraphics[width=1in,height=1.25in,clip,keepaspectratio]{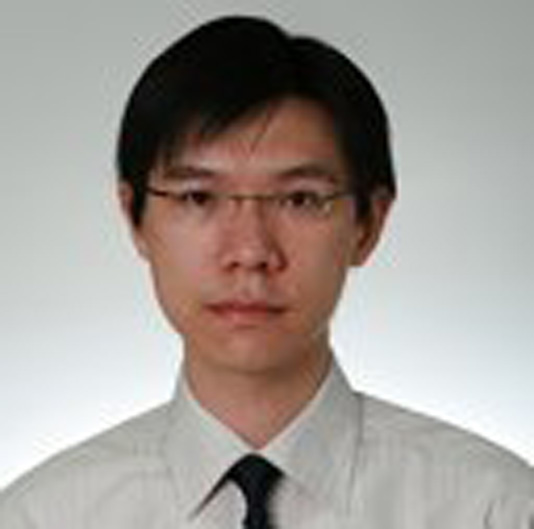}}]{Ivor W. Tsang}
	is an ARC Future Fellow and Professor of Artificial Intelligence, at University of Technology Sydney (UTS). He is also the Research Director of the UTS Flagship Research Centre for Artificial Intelligence (CAI) with more than 30 faculty members and 100 PhD students. His research focuses on transfer learning, feature selection, crowd intelligence, big data analytics for data with extremely high dimensions in features, samples and labels. He has more than 180 research papers published in top-tier journal and conference papers. According to Google Scholar, he has more than 12,000 citations and his H-index is 53. In 2009, Prof Tsang was conferred the 2008 Natural Science Award (Class II) by Ministry of Education, China, which recognized his contributions to kernel methods. In 2013, Prof Tsang received his prestigious Australian Research Council Future Fellowship for his research regarding Machine Learning on Big Data. In addition, he had received the prestigious IEEE Transactions on Neural Networks Outstanding 2004 Paper Award in 2007, the 2014 IEEE Transactions on Multimedia Prize Paper Award, and a number of best paper awards and honors from reputable international conferences, including the Best Student Paper Award at CVPR 2010. He serves as an Associate Editor for the IEEE Transactions on Big Data, the IEEE Transactions on Emerging Topics in Computational Intelligence and Neurocomputing. He is serving as a Guest Editor for the special issue of "Structured Multi-output Learning: Modelling, Algorithm, Theory and Applications" in the IEEE Transactions on Neural Networks and Learning Systems. He serves as an Area Chair/Senior PC for NeurIPS, AAAI and IJCAI. 
\end{IEEEbiography}
\vspace{-15 mm} 
\begin{IEEEbiography}[{\includegraphics[width=1in,height=1.25in,clip,keepaspectratio]{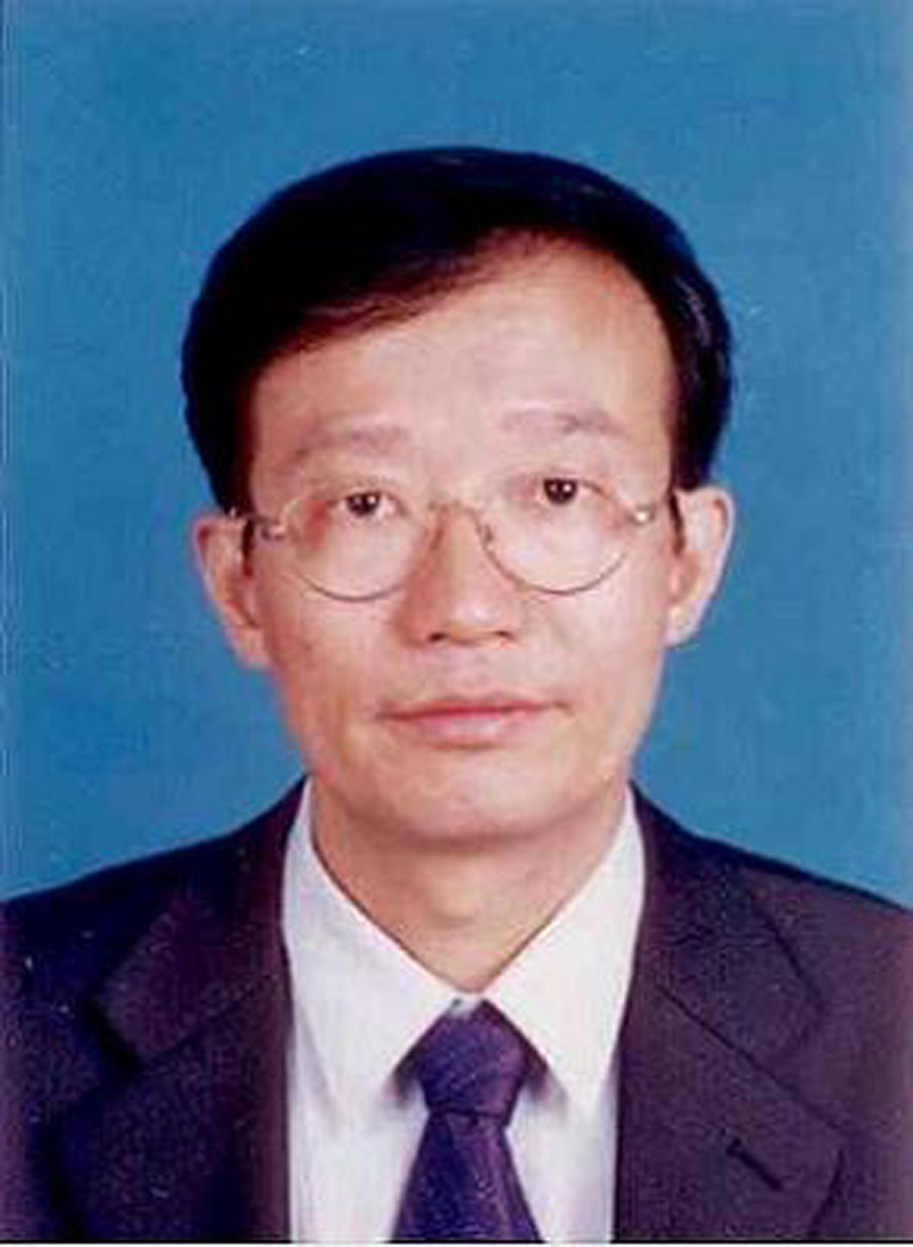}}]{Chuancai Liu}
	is currently a Full time Professor with the School of Computer Science and Engineering at Nanjing University of Science and Technology. He received the PhD degree from the China Ship Research and Development Academy in 1997. His research interests include image processing, pattern recognition and computer vision.
\end{IEEEbiography}

\end{document}